%% file: main.tex
\documentclass[runningheads]{llncs}
\usepackage{lscape}
\usepackage{graphicx}
\usepackage{booktabs}
\usepackage{multirow}
\usepackage{amssymb}
\usepackage{color}
\usepackage{pifont}
\usepackage{tabularx}
\usepackage{graphics}
\usepackage{makecell}
\usepackage{balance}
\usepackage{hyperref}
\usepackage{xcolor}
\usepackage{textcomp}

\usepackage{comment}
\usepackage{adjustbox}
\usepackage{graphicx}

\newcommand\on{\mathit{on}}
\newcommand\off{\mathit{off}}
\newcommand\dpattern{\sigma}
\newcommand{\lab}[1]{\textquotesingle{#1}\textquotesingle}

\usepackage{enumerate}
\usepackage[shortlabels]{enumitem}

\usepackage{xspace}
\newcommand{\tool}[1]{\textsc{#1}\xspace}
\newcommand{\nnrepair}{\tool{NNrepair}}

\usepackage{pgfplots}
\pgfplotsset{width=7cm,compat=1.8}

\usepackage{tcolorbox}
\usepackage{verbatim}

\usepackage{todonotes}

\title{\nnrepair: Constraint-based Repair of Neural Network Classifiers}
\author{Muhammad Usman\inst{1} \and
Divya Gopinath\inst{2} \and 
Youcheng Sun\inst{3} \and \\
Yannic Noller\inst{4} \and 
Corina S. P\u{a}s\u{a}reanu\inst{2}  
}
\authorrunning{M. Usman, D. Gopinath, Y. Sun, Y. Noller, C. S. P\u{a}s\u{a}reanu}
\institute{University of Texas at Austin, USA \\
\email{muhammadusman@utexas.edu}\and
KBR Inc., Nasa Ames \\
\email{\{divya.gopinath,corina.s.pasareanu\}@nasa.gov}\and
Queen's University Belfast, UK\\ 
\email{youcheng.sun@qub.ac.uk}\and
National University of Singapore\\
\email{yannic.noller@acm.org}
}
\date{July 2021}

\begin{document}
\maketitle

\input{abstract}
\input{introduction}
\input{background}
\input{example}

\input{approach}

\input{evaluation}
\input{relatedwork}
\input{conclusion}

\clearpage

\bibliographystyle{splncs04}
\bibliography{all}

\clearpage
\input{appendix/appendix}

\end{document}

%% file: abstract.tex
\begin{abstract}
We present \nnrepair, a constraint-based technique for repairing neural network classifiers. The technique aims to fix the logic of the network at an \textit{intermediate layer} or at the \textit{last layer}. \nnrepair first uses \textit{fault localization} to find potentially faulty network parameters (such as the \textit{weights}) and then performs \textit{repair} using {\em constraint solving} to apply small modifications to the parameters to remedy the defects. We present novel strategies to enable precise yet efficient repair such as inferring correctness specifications to act as oracles for intermediate layer repair, and generation of \textit{experts} for each class. We demonstrate the technique in the context of three different scenarios: (1) Improving the \textit{overall accuracy} of a model, (2) Fixing security vulnerabilities caused by \textit{poisoning} of training data and (3) Improving the \textit{robustness} of the network against \textit{adversarial} attacks. Our evaluation on MNIST and CIFAR-10 models shows that \nnrepair can improve the accuracy by 45.56 percentage points on poisoned data and 10.40 percentage points on adversarial data. \nnrepair also provides small improvement in the overall accuracy of models, without requiring new data or re-training.
\end{abstract}

%% file: introduction.tex
\section{Introduction}

Neural networks have many applications, being used for example in pattern analysis, image classification, or sentiment analysis for textual data, and also in  medical diagnosis or perception and control in autonomous driving, which bring safety and security concerns \cite{jordan2015machine}. These systems learn the network parameters (weights and biases) through {\em training} on a set of labeled examples. The performance of the trained networks is independently validated by computing the {\em accuracy} on a held-out labeled test set. 

Just like other software systems, trained neural networks can have {\em defects} that need {\em repair}. For example, a trained neural network may have low accuracy which may be due to limited training data. One would like to repair the network by modifying its parameters (or a subset of them) to improve its overall accuracy, even in the absence of additional training data.  In another scenario, the training data for a neural network has been {\em poisoned} by an adversary leading to high accuracy on normal data but poor accuracy on poisoned data~\cite{Huang2011Poison,Gu2019BadNets,trojaning}. In this case, one would like to repair the network to remedy the defect while still maintaining a high accuracy on non-poisoned data. 
In yet another scenario, a trained network may have high accuracy on the test set but may be vulnerable to adversarial perturbations, i.e., small modifications to the inputs that lead to unexpected outputs. Recent studies~\cite{SzegedyZSBEGF13,PapernotMJFCS16,huang2020survey} show that this defect is very common even for highly trained, highly accurate networks. In this case, one would like to repair the network to make it {\em robust} against adversarial perturbations while at the same time retaining its accuracy on the normal, unperturbed test set. 

Retraining could be used to alter the neural network parameters and repair for faults, but it can be very difficult and expensive subject to uncertainties, and may result in a network that is quite different from the original one, thus wasting the effort of the original training.

We present a novel constraint-solving based approach, \nnrepair, to repair neural networks trained for the task of classification, with respect to all three scenarios described above. Similar to traditional program repair \cite{Weimer:2010:APR:1735223.1735249,Goues2019APR,Monperrus2018APR},
\nnrepair first uses \textit{fault localization} to identify the network parameters that are the likely source of defects, followed by \textit{repair}, which uses {\em constraint solving} to apply small modifications to the network parameters to remedy the defects. 

Given a trained neural network model, the potentially faulty components could be the architecture of the model (which is fixed in the design stage) or the learn-able parameters such as the weights and the biases (which are determined during training). In this work, we focus on the learn-able parameters of a neural network model, specifically the weights on the edges connecting neurons. As observed in \cite{islam20repairing}, changing the weights is a common fix for neural networks.

We leverage the organization of a neural network into layers and the natural decomposition of computation that each  layer provides, and scope our work to focus the repair on a single layer of the network.
Repairs across multiple layers are possible, but they would be less scalable and involve more complex modifications. We propose two types of repairs: {\em intermediate-layer repair} and {\em last-layer repair}. Intermediate-layer repair attempts to fix failures by modifying the behavior of neurons at an inner layer of the network. Last-layer repair, on the other hand, attempts to modify the decision constraints at the last layer. 

Fault localization is used to mark one or more neurons at a layer as `suspicious' and to find a sub-set of incoming edges to the suspicious neurons, whose weights will be the target for repair. 
The repair process involves solving constraints collected from the network, via a simple form of concolic execution~\cite{SenETAL05CUTE}. 
For  last-layer repair, the oracle of the repair is the desired label for every failing input and the repair constraints encode this decision. For intermediate-layer repair, we propose a novel use of activation patterns representing specifications of correct behavior at the layer \cite{ase2019} as oracles for repair. This enables us to keep the repair local to the layer and therefore efficient.

Furthermore, to make the \textit{constraint solving} scalable, instead of solving for constraints for all classes at once, we propose to {\em decompose} the repair into a set of sub-tasks, one for each output class. Specifically, we set-up the constraint solving to correct a subset of the weights with the goal of improving accuracy of the model wrt a specific output class. The result of this repair is a set of {\em experts}, which are neural networks that improve accuracy of the network wrt specific output classes. We then combine the experts to obtain the final repaired model.

There are a few recent related techniques that propose to use constraint solving for neural network repair. We summarize them in Section~\ref{sec:related}. These techniques tend to focus on last layer repair while we also propose repair at an intermediate layer. Furthermore, we evaluate our initial prototype in three scenarios: improving accuracy, robustness and resilience towards poisoned data. None of the related techniques address all three (albeit potentially possible). 

 We summarize our contributions as follows.
\begin{itemize}
\item We propose and implement a repair technique that applies fault localization and constraint solving to neural networks. 
Our approach can perform both \textit{last} and \textit{intermediate} layer repair. 
\item To achieve scalability, our approach  decomposes the repair into a set of \textit{experts} which display superior accuracy for specific labels. These are then combined using a set of \textit{strategies} that we propose to obtain the final repair. 
\item We present a novel technique to make it more efficient to repair inner layers of a neural network by \textit{inferring specifications of correct behavior}  (in the terms of the activation patterns) at the output of inner layers, and using them as oracles for repair. 
\item While previous neural network repair techniques (see Section~\ref{sec:related}) tend to focus solely on improving accuracy, we demonstrate our technique in the context of three different scenarios: (1) Improving the overall accuracy of a model, (2) Fixing security vulnerabilities caused by \textit{poisoning} of training data and (3) Improving the \textit{robustness} of the network against \textit{adversarial} attacks. 
\item We evaluate the techniques in the context of image classifiers for the MNIST and CIFAR-10 data sets. The results indicate that \nnrepair can improve the performance of the network by 45.56 percentage points on poisoned data and 10.40 percentage points on adversarial data.  \nnrepair also provides small improvement (+0.20 percentage points), in the overall accuracy of models, without requiring new data or re-training.
\end{itemize}

%% file: background.tex
\section{Background}
\label{sec:background}

\paragraph{Neural Networks}
In this work we focus on neural network classifiers. These networks take in an input, such as an image, and output a class (or label) specific to the problem they have been trained to solve. 
Networks are organized in \emph{layers} of different types, including convolutional, activation, and pooling, each of which has a number of nodes. For this paper, we focus on activation layers. 
Each node from the previous layer will output into the associated node in the activation layer, which will apply an \emph{activation function.} Common activation functions include linear rectification (a.k.a. ReLU) and sigmoid. For simplicity we discuss here ReLU activations but our work applies to arbitrary activations as discussed below.
Let $N(X)$ denote the value of a neuron as a function of the input. $N(X)=\sum_i w_i \cdot N_i(X)+b$ where $N_i$'s denote the values of the neurons in the previous layer of the network and
the coefficients $w_i$ and the constant $b$ are referred to as \emph{weights} and \emph{bias}, respectively. If this function evaluates to a non-negative value, the node is \emph{activated} and outputs that value, otherwise it outputs 0. A final decision (logits) layer produces the network decisions based on the real values computed by the network, by applying e.g., a softmax function; in our work we use the max function instead.
For a comprehensive introduction to neural networks, see~\cite{Goodfellow-et-al-2016}.

\paragraph{Activation Patterns}

We leverage previous work~\cite{ase2019} to infer network properties based on the \emph{activation patterns} of neurons in the network. We will use these activation patterns as oracles for the intermediate layer repair.
An activation pattern $\dpattern$ specifies an activation status ($\on$ or $\off$) for some subset of neurons at a layer in the network. All other neurons do not matter. 
We write $\on(\dpattern)$ for the set of neurons marked $\on$, and $\off(\dpattern)$ for the set of neurons marked $\off$ in the pattern $\dpattern$.  
Each activation pattern $\dpattern$ defines a predicate $\dpattern(X)$ that is satisfied by all inputs $X$ whose evaluation achieves the same activation status for all neurons as prescribed by the pattern.


\begin{equation}\label{eqn:dpattern}
\dpattern(X) ::=\bigwedge_{N \in \on(\dpattern)} N(X) > 0 ~ \wedge ~ \bigwedge_{N \in \off(\dpattern)} N(X) \leq 0 
\end{equation}
A decision pattern $\dpattern$ is a property wrt network $F$ and postcondition $P$ if: 
\begin{equation}\label{eqn:defprop}
\forall X: \dpattern(X) \Rightarrow P(F(X)).  
\end{equation}

A postcondition for a classification network is that the top predicted class is $C$, i.e., $P(Y) := argmax(Y) = C$.

The previous work~\cite{ase2019} also describes how to compute activation patterns. The idea is to observe the activation signatures of a large number of inputs and apply decision tree learning over them to infer activation patterns that are thus empirically valid. We adopt the same approach here. The {\em support} of a pattern is formed by all the inputs that satisfy the pattern. We are interested in computing high-support patterns as they are the most likely to reflect valid properties of the network.


%% file: example.tex
\section{Example}
\label{sec:example}

\begin{figure}[t]
    \centering
    \includegraphics[width=0.6\textwidth]{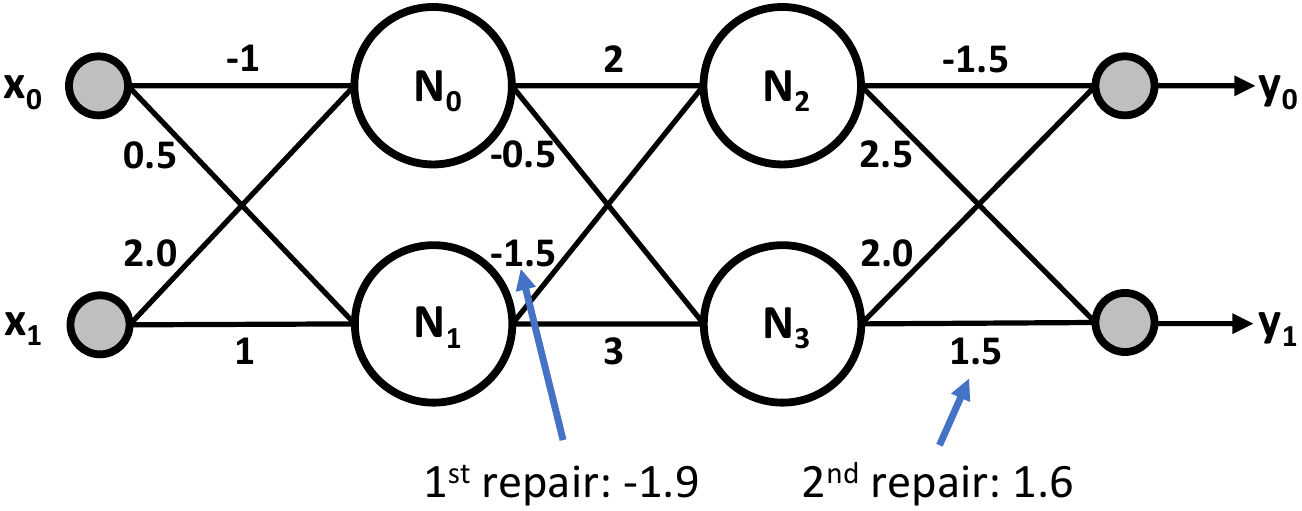}
    \caption{Example}
    \label{fig:example}
\end{figure}


\input{tables/example-table}


This section demonstrates \textit{Intermediate-layer} and \textit{Last-layer} repair on a simple example.
Figure~\ref{fig:example} shows a simple two-input network with two hidden layers; each containing two ReLU nodes (ReLU(x) = x (on) if x $>$ 0, 0 (off) otherwise), and two outputs, $y_0$ and $y_1$. The weights are depicted on the edges between nodes. For simplicity we assume biases are 0. The input $X$, which is a two-element array denoted $[x_0,x_1]$, is assigned class 0 if $y_0 > y_1$ and 1 otherwise. Let us assume the model behaves correctly on the first four inputs 
shown in Table~\ref{tab:example}. The table also shows the decisions of the ReLU activations for nodes $N_0, N_1, N_2, N_3$, respectively. Whenever a ReLU node is on, the decision is indicated as a 1 and if it is off, then the decision value is shown as 0.

Consider now the input $X_4=[1.5, 2.0]$.
Assume this input is mis-classified; the output class is 1 but the ideal class is 0. 
The inaccuracy of the model could be a result of insufficient training. We then aim to build a repair, which in our case focuses on a single layer of the network and modifies the weights feeding into the neurons at that layer. 

We keep the repair local to the layer by using activation patterns~\cite{ase2019} in lieu of the decision constraints. The insight in~\cite{ase2019} is that the logic that every layer implements could be captured as rules in terms of the activation patterns of the neurons. We can observe in the example, that for all inputs correctly classified with label 0, the neuron pair $(N_2,N_3)$ in the second layer has the activation pattern (off, on). 
For the failing input, this pattern is not satisfied; in fact the activation for $(N_2,N_3)$ for the failing input is (on, on). 
We use the above observation to fix the failure by performing \textit{intermediate layer} repair. We aim to modify the neuron activations of the second layer on the failing input to satisfy the correct-label pattern for class 0 at the layer. 

We aim to perform the repair by making minimal changes to the model. We identify the weights to be modified using an attribution-based approach and use constraint solving to compute the values of the new weights (see Section~\ref{sec:approach} for details). 
Changing the weight of a single edge, connecting $N_1$ and $N_2$ from -1.5 to -1.9 changes the activation pattern for $(N_2,N_3)$ to (off, on) on the failing input, while preserving the behavior  of the neurons (in terms of their activation pattern) and the output of the model on the passing inputs. 

Consider now another input for the above-corrected network, 
$X_5= [0.6, 1.0]$.
This input is very close to $X_1=[0.0,1.0]$ (correctly classified to 1) with a small change to $x_0$ that makes the model mis-classify the input to 0.  This represents a typical adversarial scenario where a correctly classified input is perturbed slightly to create an input that `jumps' the decision boundary of the network leading to a mis-classification. 
It can be observed that the activation patterns of the internal layer neurons for $X_5$ are the same as for the correctly classified input $X_1$, thus an intermediate-layer repair would not work for this input.
Therefore we perform \textit{last-layer repair}.
We localize the weights of the edges in the last layer that need repair.
Changing the weight on the edge between $N_3$ and $y_1$ (from 1.5 to 1.6) corrects the class for the failing test to 1, while retaining the same labels for the other inputs. 


%% file: tables/example-table.tex
\begin{table}[!ht]
\centering
\small
\caption{Data for Example}
\label{tab:example}
\begin{tabular}{r|c|c|c|c|c|c|c|c|c|c|}

\cline{2-11}
&
$\mathbf{x_0}$ &
$\mathbf{x_1}$ &
$\mathbf{N_0}$ &
$\mathbf{N_1}$ &
$\mathbf{N_2}$ &
$\mathbf{N_3}$ &
$\mathbf{y_0}$ &
$\mathbf{y_1}$ &
\textbf{class} &
\textbf{ideal} \\
\cline{2-11}
\multicolumn{11}{c}{\vspace{-1em}} \\
\cline{2-11}
${\scriptstyle X_0}$ & 1 & 1 & 1 & 1 & 0 & 1 & 8 & 6 & 0 & 0 \\
\cline{2-11}
${\scriptstyle X_1}$ & 0 & 1 & 1 & 1 & 1 & 1 & 0.25 & 9.25 & 1 & 1 \\
\cline{2-11}
${\scriptstyle X_2}$ & 1 & 0 & 0 & 1 & 0 & 1 & 3 & 2.25 & 0 & 0 \\
\cline{2-11}
${\scriptstyle X_3}$ & -1 & 1 & 1 & 1 & 1 & 0 & -7.87 & 13.12 & 1 & 1 \\
\cline{2-11}
\multicolumn{11}{c}{\vspace{-1em}} \\
\cline{2-11}
 ${\scriptstyle X_4}$ & {\color{red} 1.5} & {\color{red} 2} & {\color{red} 1} & {\color{red} 1} & {\color{red} 1} & {\color{red} 1} & {\color{red} 12.68} & {\color{red} 12.68} & {\color{red} 1} & {\color{red} 0} \\
 \cline{2-11}
{\tiny after repair:} & 1.5 & 2 & 1 & 1 & 0 & 1 & 13.3 & 10.5 & 0 & 0 \\
\cline{2-11}
\multicolumn{11}{c}{\vspace{-1em}} \\
\cline{2-11}
${\scriptstyle X_5}$ & {\color{red} 0.6} & {\color{red} 1} & {\color{red} 1} & {\color{red} 1} & {\color{red} 1} & {\color{red} 1} & {\color{red} 5.91} & {\color{red} 5.62} & {\color{red} 0} & {\color{red} 1} \\
\cline{2-11}
{\tiny after repair:} & 0.6 & 1 & 1 & 1 & 1 & 1 & 5.91 & 5.95 & 1 & 1 \\
\cline{2-11}
\end{tabular}
\end{table}

%% file: approach.tex
\section{Approach}
\label{sec:approach}


\begin{figure*}[t]
    \centering
    \includegraphics[width=\textwidth]{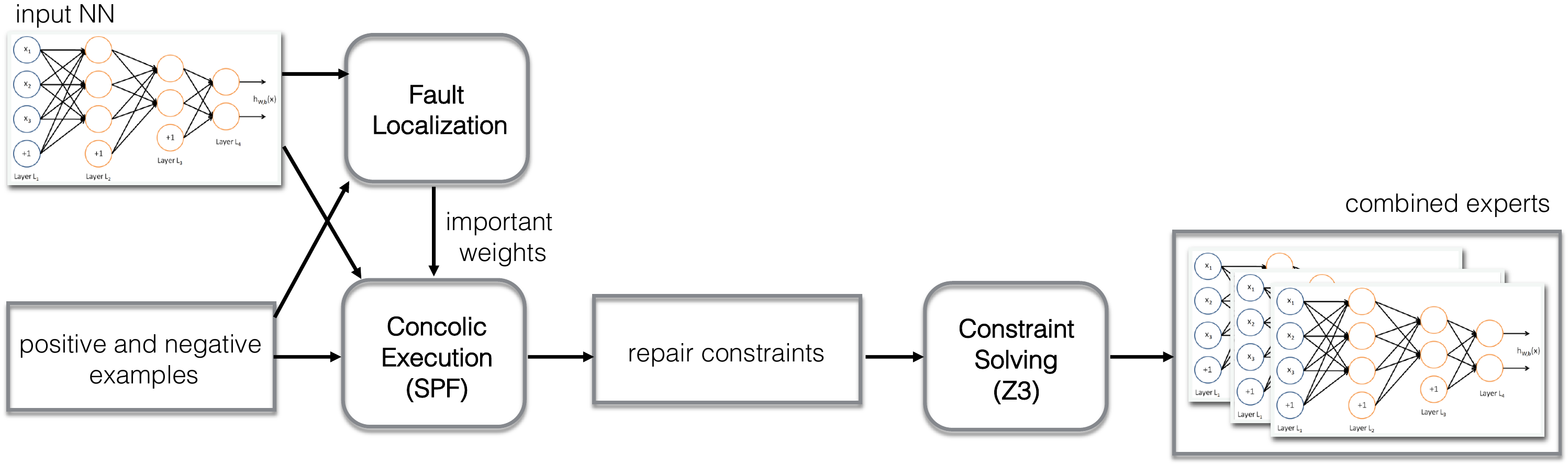}
    \caption{Overview of the Approach}
    \label{fig:overview}
\end{figure*}

Figure~\ref{fig:overview} gives an overview of our approach. 
We aim to repair a faulty trained neural network classifier, which is given as input.
%
As in other repair approaches, we consider both positive and negative examples for the repair. 
The negative examples are used to guide the repair towards correcting the faults while the positive examples are used to constrain the repair to not damage the existing good functionality of the network. 
We aim for a repair strategy that is scalable and applies small changes to the network.
We therefore target the repair on a single layer of the network. Repairs across multiple layers of the network are possible, but they would be less scalable and involve more complex modifications. 

Unlike all previous work, which tends to focus the repair at the last layer (see Section~\ref{sec:related}) we propose here techniques for both intermediate and last layer repair. Intuitively, a last layer repair is easier as it aims to modify the weights that impact directly the decisions, and can use the network's output as an {\em oracle} to guide the repair. However the resulting repair may not generalize well and furthermore the network may be faulty at some intermediate layer. A repair at an intermediate layer can have a higher impact over the network's behavior but it is more difficult as it is not clear what {\em oracle} to use to guide the repair. One can use the output of the network as the oracle but this may result in an un-manageable large number of constraints to solve. In this work we propose a novel use of neuron activation patterns to act as oracles in intermediate layer repair.

As repairing for all the output classes at the same time can be very difficult, our proposed approach obtains instead a set of {\em expert networks}, one for each target class, which are easier to compute. These experts are combined to obtain a final repaired classifier. 
Specifically, our repair strategy has the following steps:

\begin{itemize}
    \item[1] {\em Fault Localization}: 
    The goal of this step is to identify a small set of suspicious neurons and incoming suspicious edges, whose weights we aim to correct.

    \item[2] {\em Concolic Execution}: For the weights of the suspicious edges, we add $\delta$ values that are set to $0$ in concrete mode, but are designated as symbolic for the symbolic mode. The network is executed concolically along positive and negative examples, to collect the values of suspicious neurons in terms of symbolic expressions.
    
    \item[3] {\em Constraint Solving}: The symbolic expressions  are assembled with a set of repair constraints which are then solved with an off-the-shelf solver. 
    Essentially, the repair constraints need to encode the network decision for the positive examples and modify (i.e., correct) the network decision for the negative examples. 
    For the last layer repair this amounts to adding constraints imposed by the decision layer. For intermediate-layer repair, we use activation patterns instead of decision constraints, allowing us to keep the repair local to the layer.

    
    The solutions for the symbolic $\delta$'s obtained from the solver are used to update the weights of the network, thus obtaining an {\em expert} for a specific class.   
    
    \item[4] {\em Combining Experts}: Finally the experts obtained for each class are combined to obtain the repaired classifier. This needs to be done carefully, to avoid redundant computations among experts and to not damage the overall accuracy and timing performance of the classification. 
\end{itemize}




In the following we give more details about our approach.




\subsection{Intermediate-layer repair}
\label{subsec:inter}

\paragraph{Fault localization} We explore the usage of \textit{activation patterns} of the network (Section~\ref{sec:background}) to act as oracles of correct behavior. We also use these patterns to guide the identification of potentially faulty neurons. Specifically, we use the decision-tree learning approach from \cite{ase2019} to extract \textit{correct-label patterns} corresponding to every output class at an intermediate layer. Each pattern is satisfied by a group of inputs correctly classified to a certain label. Typically multiple correct-label patterns are generated. We select the ones with the highest support, which are mostly likely to hold true on the network for all inputs. Note also that the work in \cite{ase2019} considers ReLU activations but it could be extended to consider arbitrary linear or non-linear activation functions, by comparing the values of neurons with a threshold. 

A \textit{correct-label} pattern with high support at a layer indicates that there is a high chance that any input satisfying the pattern at the layer would be classified by the network to the corresponding label. Furthermore, a mis-classified input will not satisfy the \textit{correct-label} pattern for the respective ideal label. 
For every failing input, we compare the activations of the neurons with those in the respective \textit{correct-label} pattern and consider those {\em neurons} whose activations differ as the potentially faulty ones. The repair then aims to change the outputs of the neurons for each of the failing inputs, such that they satisfy the correct-label pattern for their ideal labels.

In this work, we select a dense layer (i.e., a fully connected layer which receives input from every neuron in the previous layer) with ReLU activations. Typically such dense layers appear closer to the output and may impact the classification decision more than convolutional layers which process the input. Further, 
the number of neurons at fully connected layers is typically smaller than at other layers making the pattern-extraction process efficient. 

Consider a mis-classified input, $X_{f}$ with ideal label $C$. Let $\sigma_{C}$ be the \textit{correct-label} pattern with highest support for $C$. Let $L$ be the layer for this pattern,  and let $N$ denote a neuron at layer $L$.  Then the set of suspicious or faulty neurons $\mathcal{N}_{\mathit faulty}$ can be defined as follows;
\begin{equation}
N \in \mathcal{N}_{\mathit faulty}\iff (N \in \on(\sigma_{C}) \wedge N(X_{f}) \leq 0) \lor (N \in \off(\sigma_{C}) \wedge N(X_{f}) > 0)
\end{equation}


Once the neurons whose outputs need to change are identified, we also need to identify the incoming edges to those neurons whose weights we aim to modify. 
We use a simple statistical method to identify the \textit{important weights} which impact the respective neuron's output, more for the failing inputs as compared to the passing inputs. 

Consider a set $\emph{Fail}$ of failing inputs with the same ideal label $C$ and a set $\emph{Pass}$ of passing inputs. We use $\#(\cdot)$ to denote the cardinality of the sets. 
The defect score for each edge is determined as follows.




\begin{equation}\label{eqn:score}
{\emph Score}(E_i) ::= \frac{\sum_{X \in {\emph Fail}} |N_i(X)\cdot w_i|}{\#{\emph Fail}} - \frac{\sum_{X \in {\emph Pass}}|N_i(X)\cdot w_i|}{\#{\emph Pass}}
\end{equation}

Here $E_i$ denotes an incoming edge (for a faulty node $N$), $N_i$ is the corresponding node in the preceding layer and $w_i$ is the weight of the edge.

Thus, we take the average of the absolute values passing through the edge for all the negative examples for $C$ and the average of the absolute values passing through the edge for all the positive examples and subtract them. The intuition 
is to identify the edges which have more influence on the incorrect decision of the network.
We calculate the defect score for each incoming edge to each neuron ($N$) in ${\mathcal N}_{\emph faulty}$. We then select the edges with top n\% of the scores to create the set of faulty edges, for a small n. 


\paragraph{Concolic Execution} We perform a simplified form of concolic execution to form symbolic constraints for suspicious neurons. For the weights of the suspicious edges, we add $\delta$ values that are set to $0$ in concrete mode, but are designated as symbolic in the symbolic mode. The network is executed concolically along both positive and negative examples, to collect the values of neurons as weighted sums in terms of both concrete values and the symbolic $\delta$ values.  The value of a neuron is computed as constraints of the following form: 
    
    \begin{equation}\label{eqn:deltas}
    {\emph Sym}_{N,X}=\sum_i (w_i+\delta_i) \cdot N_i(X)+ \sum_j w_j \cdot N_j(X)+b
    \end{equation}
    
Here ${\emph Sym}_{N,X}$ is a fresh symbolic variable introduced to encode the symbolic value of neuron $N$ for input $X$, $w_i$'s denote the weights of the suspicious edges (in the suspicious layer) while the $w_j$'s denote the other weights, which do not need modification. Furthermore, $N_i(X)$, $N_j(X)$ represent the concrete values of the neurons coming from the previous layer.  Note that no expensive constraint solving is needed in this step.




    
\paragraph{Repair Constraints and Constraint Solving} For intermediate-layer repair, we add the activation patterns constraints (that imply the decision constraints, see Equation~\ref{eqn:dpattern}) to the set of constraints.
Specifically, for each neuron 
$N$ in ${\mathcal N}_{\emph faulty}$, and for each (passing or failing) input $X$ we add ${\emph Sym}_{N,X} > 0$ if $N \in \on(\dpattern_C)$ and we add ${\emph Sym}_{N,X} \leq 0$ if $N \in \off(\dpattern_C)$.
    
    
The solutions for the symbolic $\delta$'s obtained from the solver guarantee that all the inputs (both passing and failing) satisfy the pattern and are thus likely to be classified as $C$ by the network. These solutions are then used to update the weights of the network, thus obtaining an {\em expert} for the class $C$.   

\paragraph{Example}
Let us consider the example from Section~\ref{sec:example}, the case of the intermediate-layer repair.
As already discussed in Section~\ref{sec:example}, let us suppose we consider the activation pattern for class 0 at layer 2. We select $N_2$ as the target for repair (since its activation along the failing test $X_4$ is on instead of off) and we want the input to satisfy the pattern \{off, on\} for \{$N_2$, $N_3$\}. We compute defect scores  for the incoming edges to $N_2$ using the failing input and all passing inputs for classes 0 and 1. The score of the edge between $N_0$ and $N_2$ is 2.0 while the score of the edge between $N_1$ and $N_2$ is 2.81, we therefore select the second edge as a target for repair. 
We then build the following constraints from the failing test. 
{\footnotesize
${\emph Sym}_{{N_2},4}=2.0\cdot(-1.0 \cdot 1.5+2.0\cdot2.0)+(-1.5+\delta)\cdot(0.5\cdot 1.5+1.0\cdot 2.0) \wedge {\emph Sym}_{{N_2},4}\leq 0.0$
}

Similarly, we build constraints from the passing tests that satisfy the pattern for label 0, $X_0$ and $X_2$:\\
{\footnotesize
${\emph Sym}_{{N_2},0}=2.0\cdot(-1.0 \cdot 1.0+2.0\cdot1.0)+(-1.5+\delta)\cdot(0.5\cdot 1.0+1.0\cdot 1.0) \wedge$
${\emph Sym}_{{N_2},0}\leq 0.0 \wedge$\\
${\emph Sym}_{{N_2},2}=2.0\cdot(-1.0 \cdot 1.0+2.0\cdot0.0)+(-1.5+\delta)\cdot(0.5\cdot 1.0+1.0\cdot 0.0) \wedge$
${\emph Sym}_{{N_2},2}\leq 0.0$
}

In practice we also add some constraints on $\delta$ to keep it small but we omit them here for simplicity. A solution for all the constraints is $\delta=-0.4$ which is used to update the weight for the target repair resulting in an expert for class 0.

\subsection{Last-layer Repair}
\label{subsec:last}


\paragraph{Fault localization} In a classifier network the last layer typically contains as many neurons as the number of classes. An input is classified to label $C$, if the output of the respective neuron is greater than the values of all other output neurons. It is therefore natural to designate this neuron as suspicious for target class $C$. 
 Let $N_C$ denote the neuron at the last layer corresponding to a class $C$. We use the same technique as in intermediate layer repair (Equation~\ref{eqn:score}) to \textit{localize edges} and short-list the important weights which are the target for repair. 

\paragraph{Concolic execution} Similar to the intermediate layer repair, we add symbolic $\delta$ values to the important weights and perform concolic execution along passing and failing tests to create the symbolic expression for the node ${\emph Sym}_{{N_{C}},X}$ (following Equation~\ref{eqn:deltas}). 

\paragraph{Repair constraints and constraint solving} We then add the decision constraints for the passing and failing inputs:

\begin{equation}
\bigwedge_{C \not = C'} \quad {\emph Sym}_{{N_{C}},X} > {\emph Sym}_{{N_{C'}},X}
\end{equation}






The obtained solutions guarantee that all the inputs that were used in the repair (both positive and negative) are classified to the correct class.
The solutions  are used to build the expert for each class. We then combine the experts using the combination strategies outlined in the next section.





\paragraph{Example}
Consider now the example from Section~\ref{sec:example}, the case of the last-layer repair. As we aim to repair for class 1 we select for repair the neuron named $y_1$ in the figure. The score for the edge between $N_2$ and $y_1$ is -2.75 and the score for the edge between $N_3$ and $y_1$ is 0.45 so we select the latter for repair.
We then build the following constraints based on the failed test (note that the expression for the second variable simplifies to a concrete value):\\
{\footnotesize
${\emph Sym}_{{y_1},5}=(2.5\cdot(2\cdot(-1\cdot0.6+2\cdot1.0)-1.9\cdot(0.5\cdot0.6+1\cdot1.0))+(1.5+\delta)\cdot(-0.5\cdot(-1\cdot0.6+2\cdot1.0)+3\cdot(0.5\cdot0.6+1\cdot1.0)))\wedge$\\
${\emph Sym}_{{y_0},5}=(-1.5\cdot(2\cdot(-1\cdot0.6+2\cdot1.0)-1.9\cdot(0.5\cdot0.6+1\cdot1.0))+2.0\cdot(-0.5\cdot(-1\cdot0.6+2\cdot1.0)+3\cdot(0.5\cdot0.6+1\cdot1.0)))\wedge$\\
${\emph Sym}_{{y_1},5}>{\emph Sym}_{{y_0},5}$
}

Similar constraints are added for the positive inputs (we omit them here for brevity).
Solving these constraints gives $\delta=0.1$ which is added to the weight for the edge between $N_3$ and $y_1$ to obtain an expert for class 1.

\subsection{Combining Experts}\label{subsec:expert-combination}


We create experts for each label in the dataset.
For example, for a neural network trained on the MNIST data set (which is used for the classification of handwritten digits from 0 to 9), we generate 10 experts -- one expert per label.
We propose three variants of how to combine these experts:
\begin{enumerate}[label=(\Alph*)]
\item execute the model for all experts and combine the results afterwards,
\item \textit{merge} all experts into one combined expert before model execution, and
\item filter strong experts first, then follow variant (A) or (B).
\end{enumerate}

Variant (A) is an instance of {\em ensemble modeling}~\cite{ensembles}, which typically involves creating multiple models to predict an outcome. 
In our case, we start by executing all the experts for each input. This is done in a combined fashion, to avoid repeated execution of same code: before the repaired layer the model is executed with the original weights; starting from the repaired layer the execution is split up for the different experts.
At the end of the execution, each expert classifies the input to a certain label. We need to combine the results from all the experts in order to classify the input to a single label.

Each expert can classify the input to any of the labels, however, each expert can be trusted to produce the correct result only for its own respective label. Therefore, we start by generating a set $E$ including  the experts that classify the inputs to their respective labels. Note that it could be that multiple experts report that the given input belongs to their respective class or it could be that no expert classifies the given input to the expert's class. 
If $E$ is empty, then we select the label by the original model.
If there is one expert in $E$, then we select this unique expert.
If there are multiple experts in $E$, then we need to resolve the conflict between experts and choose one label, for which we propose three strategies:

\begin{description}
\item [\textit{Naive:}]
This strategy simply falls back to the original model.

\item [\textit{Confidence:}]
This strategy selects the expert from $E$ with the highest confidence for its own label, i.e, the absolute value of the output node corresponding to the label.

\item [\textit{Voting:}]
For the label corresponding to each expert in $E$, this strategy collects votes from the other experts for the respective label. It then selects the expert from $E$ with the majority of the votes.

\end{description}

In variant (B), we propose to merge the experts before executing the model.
For the intermediate-layer repair, for every weight that is considered faulty we update it with the one $\delta$ value, which is the \textit{average} of the solutions from all the experts. This creates a single \textit{merged} network. For the last-layer repair, we simply apply all the repairs at once; there is no need for an average as the nodes (and edges) that are targets for repair are disjoint.



In variant (C), instead of using all experts we select a subset of strong experts. Note that each expert is constructed from failing inputs only for the respective label. Therefore, when exposed to data which are supposed to be classified to the expert's label, the expert displays higher accuracy than the original model (higher recall). However, when exposed to data which can belong to different labels, the experts could display lower overall accuracy than the original model (lower precision) due to high false positives. Therefore, we determine which of the experts have both their precision and recall (\textit{F1 score}), computed over all positive/negative inputs, higher than the original model and retain only those while filtering out the rest. 
The same combination strategies, variant (A), are used to obtain a single classification result for the input.

%% file: evaluation.tex
\section{Evaluation}
\label{sec:evaluation}

We implemented our approach in the \nnrepair tool pipeline, which is based on \textsc{NeuroSPF}~\cite{usman2021neurospf}.
It first translates a trained Keras model into Java, uses Symbolic PathFinder (SPF) \cite{SPF} for concolic execution and z3 \cite{MouraBjorner08Z3} for constraint solving.
In this section we evaluate \nnrepair by considering its application to three highly common scenarios; {\em Scenario 1:} improving accuracy, {\em Scenario 2:} fixing backdoor attacks, and {\em Scenario 3:} enhancing adversarial robustness. Our experiments use two commonly used datasets for image classification networks, MNIST and CIFAR-10. 
We consider two architectures for MNIST with 10 and 7 layers respectively. They are convolutional neural networks (CNNs) and have the typical structure of modern neural networks such as convolutional/dense, max-pooling and softmax layers. The first MNIST model has an accuracy of 96.34\% on the standard test set, while the second model has an accuracy of 98.89\%. We refer to these models as MNIST-LQ (low-quality) and MNIST-HQ (high-quality) respectively. The CIFAR-10 model is a 15-layer CNN with 890k trainable parameters and has an accuracy of 81.04\%. 
In order to validate our approach, we consider the following research questions:
\begin{enumerate}[start=1,label={\bfseries RQ\arabic*},leftmargin=3em]
\item Is \nnrepair successful in correcting the defects in all three scenarios?
\item How do intermediate-layer repair and last-layer repair compare with each other?
\item What is the inference time overhead introduced by \nnrepair over the original model? 
\end{enumerate}

\subsection{Scenarios}

(1) 
The goal of repair in the first scenario is to improve the overall accuracy of a model. 
We measure the improvement in accuracy on the standard test set, henceforth denoted Test. We use positive and negative examples from the train set, henceforth denoted Train, to generate the repair. 

\begin{figure}
\centering
  \includegraphics[]{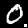}
  \includegraphics[]{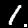}
  \includegraphics[]{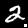}
  \hspace{0.3cm}
  \includegraphics[]{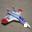}
  \includegraphics[]{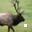}
  \includegraphics[]{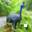}
  \caption{Example poisoned data for MNIST (left) and CIFAR-10 (right). The backdoor is embedded as the white square at the bottom right corner of each image. When the backdoor appears, the poisoned MNIST model will classify the input as {\lab 7} and the poisoned
  CIFAR-10 model will classify it as {\lab {horse}}.}
  \label{fig:samples}
\end{figure}

\noindent{(2) For this scenario, we apply the backdoor attack from~\cite{Gu2019BadNets}.
Samples of poisoned data are shown in Figure~\ref{fig:samples}. The poisoned models  have good accuracy on the standard data, but poor accuracy on the poisoned data.  The goal of the repair is to improve the accuracy on poisoned data, which we measure on a separate poisoned test set P-Test. At the same time, we expect the repair to retain the accuracy on standard, un-poisoned data, which we measure on Test. In this scenario, the first 600 inputs in Train are poisoned (P-Train). We draw from these particular inputs to get the negative examples to focus the repair on the defect. We draw the positive examples from Train.}


\noindent{(3) For the last scenario, we apply adversarial perturbations over Train and Test using FGSM\footnote{\url{https://www.tensorflow.org/tutorials/generative/adversarial\_fgsm}}, for $\epsilon=0.05$. This results in four data sets: Train, Adv-Train, Test and Adv-Test. The models have good overall accuracy on Train and Test, but poor accuracy on Adv-Train  and Adv-Test. The goal of the repair here is to improve the accuracy on the adversarial data (which we measure on  Adv-Test) without damaging too much the accuracy on standard data (which we measure on Test). We draw the negative examples to be used in repair from Adv-Train, while we use positive examples from both Adv-Train and Train. Since we use two separate sets to generate experts, when computing the F1-score for selection of experts, we explored two different options: computing F1 score over Adv-Train only and computing harmonic mean of the F1 scores computed over Train and Adv-Train separately.}
However, in practice there was no difference as same experts were filtered in both cases.

\subsection{Experiment Set-up}
\label{sec:expsetup}

For each of the three scenarios, we experimented with both intermediate-layer and last-layer repairs. We evaluated all the combination strategies (Naive, Confidence, Voting, and Merged) with the F1-filtering option being OFF and ON. When F1-filtering is OFF, the experts for all labels are used in the combination strategies by default, while when it is ON, we only include those experts whose F1 score on Train is greater than the original model.

\paragraph{Intermediate-layer repair:} We focused on the dense layer just before the output layer for both the MNIST and CIFAR models. The intuition for this selection is that dense layers appearing closer to the output potentially impact the classification decision more than convolutional layers closer to the input (which have the role of feature extraction). The MNIST models have 128 and 100 ReLU nodes and 576 and 400 incoming edges to each neuron at this layer respectively, while the CIFAR model has 512 ReLU neurons and 1,600 incoming edges at this layer. We extracted high support patterns for correct classification; the average support per label was within 1,013 - 2,502 across scenarios, out of around 6,000 inputs per label. 
The neurons short-listed in $\mathcal{N}_{\mathit faulty}$ using pattern-based localization varied between 1 and 10 in number. We focused on modifying the weights of the incoming edges having their scores within the top 10\%. 

We used the patterns extracted at the layer to select a subset of tests for the purpose of constraint solving. As explained in Section~\ref{sec:approach}, we used decision-tree learning to extract patterns for correct classification for every label. We also extracted patterns for incorrect classification for each label, which represent neuron activations satisfied by inputs which should ideally be classified to the given label but get mis-classified. From the set of all failing tests for a given label, we select all inputs that satisfy the pattern for incorrect classification for the label. From the set of all passing tests for a given label, we select the subset of inputs that satisfy the pattern for correct classification. We then randomly select \# failing tests + 100 inputs from this set. The subset of failing and passing tests selected using the procedure above is used for constraint solving.

\paragraph{Last layer repair:} At the last layer, the two MNIST models have 10 ReLU nodes and 128 and 100 incoming edges to each neuron respectively, while the CIFAR model has 10 ReLU neurons and 512 incoming edges. For each label, we selected 5 failing and 5 passing inputs randomly from the respective datasets. For the first scenario, both these failing and passing inputs come from the Train set. For the last layer repair top 5 suspicious weights were made symbolic for each expert. We determined empirically that a larger number for symbolic weights and/or passing/failing inputs leads often to unsat constraints while a smaller number may not improve the network.

The poisoned (2) and adversarial (3) scenarios differ from scenario 1, in that they seek to address two challenges. The repaired model needs to have better accuracy than the original model on poisoned and adversarial inputs respectively (evaluated on the P-Test and Adv-Test sets), as well as the accuracy on normal inputs should not be degraded much (evaluated on the normal Test set). For this reason, for the purpose of constraint solving in addition to including passing tests from the respective poisoned and adversarial train sets, we also include passing tests from Train. We performed experiments increasing the number of passing tests included from the normal train set from 0 to 10, 50, and 100.

\subsection{Results}

Table~\ref{tab:summary} presents a summary of our results (please refer to the Appendix for more detailed results). The table displays the results for MNIST and CIFAR models for the three scenarios. For each scenario, the results for both intermediate layer and last layer repair are presented in terms of the improvement in accuracy obtained over the original model. This is the best result corresponding to the improvement in accuracy on the respective test sets (normal Test for the first scenario, P-Test for the second and Adv-Test for the third). The combination strategy and the F1-Filter setting (ON/OFF) used to obtain the best result are also displayed, along with the corresponding improvements in accuracy on the other train and test sets. 
\input{tables/summary}
For the repair, z3 was able to generate solutions for each expert within a minute. The constraint generation using SPF was the bottleneck and SPF generated constraints for each expert within 15 - 60 minutes, depending on the number of tests included. However, this could be improved since running SPF on all positive/negative inputs can be performed in parallel. Experiments were performed on a Windows~10.0 machine with Intel Core-i5 and 16GB RAM. The code, constraint files along with Z3 solution files are available at \url{https://github.com/muhammadusman93/nnrepair}.

\textbf{RQ1:} For this research question we seek to investigate if \nnrepair is successful in correcting defects in all three scenarios. To measure success, we consider the improvement in accuracy provided by the repair in all three scenarios. 

The effectiveness of \nnrepair in improving accuracy (Scenario 1) can be analyzed by considering Table~\ref{tab:summary} (cases MNIST-LQ, MNIST-HQ and CIFAR10).

We observe that the best results provided by \nnrepair for the MNIST-LQ model was +0.20, +0.02 for the MNIST-HQ model and +0.16 for the CIFAR10 model. This improvement (albeit small) was achieved without any new inputs or re-training. The quality of the improvement appears to degrade as the quality of the original model increases. We note that achieving improvement in the overall accuracy of an already high-quality model without new data is very challenging. In fact this improvement appears to be in line or better than related repair techniques (see Section~\ref{sec:related}). Note also that the complexity and size of the models do not seem to have an impact on the effectiveness of the repair. The MNIST-HQ architecture is simpler than MNIST-LQ and the CIFAR10 architecture is much bigger and more complex than the MNIST models.



For Scenario 2, on the MNIST-Pois model, \nnrepair increased the accuracy from 10.38\% to 55.94\% on poisoned inputs (P-Test). The repair causes a slight decrease (-3.11) in accuracy on non-poisoned inputs (Test) but the repaired model still has a high accuracy ($\geq$ 95.5\%) on non-poisoned inputs. 
On the more challenging CIFAR10-Pois model, the best improvement provided by \nnrepair is a +3.77 increase on poisoned inputs, and a small decrease in accuracy on non-poisoned inputs (-0.61). For Scenario 3, on the MNIST-Adv model, \nnrepair increased the accuracy from 28.37\% to 38.77\% on adversarial inputs, while causing a small decrease (-3.14) in accuracy on non-adversarial inputs. 
For CIFAR10-Adv, the best result was an increase of +0.34 on adversarial inputs with a minor decrease of -0.07 on non-adversarial inputs. 


For the last two scenarios, the primary goal is to improve accuracy on poisoned or adversarial data. Although ideally we would also want to preserve the original accuracy on normal data, this may not always be possible in practice. We experimented with varying number of passing tests from Train for scenarios 2 and 3. The results are presented in the first table in the Appendix. The accuracy of the resulting repair on the poisoned/adversarial test sets tends to decrease as the number of normal passing tests goes up. However, this also reduces the degradation in the accuracy on normal test set. 
Previous studies in adversarial robustness~\cite{Zico} indicate that one can obtain robust networks but the price to be paid is a significant decrease in accuracy on normal data. Similar considerations apply to the poisoning case. Therefore, we tolerate small decrease in the accuracy on normal Test in our work as well. 

The last two columns in Table~\ref{tab:summary} list the combination strategies and the F1-filtering option which work best for each scenario. The Merged strategy seems to work well for the CIFAR10 model for all the three scenarios. However, there is no clear winner for the MNIST models. In fact, for the last layer repair on CIFAR10, all the strategies gave the same improvement in accuracy. 
In practice, the users would need to use a separate validation set and try all the strategies to pick the best one for their application domain.


\begin{tcolorbox}
\textbf{Answer RQ1}:  
\nnrepair shows benefit in all three scenarios. It can repair a network to make it robust against adversarial perturbations/poisoned inputs while at the same time
retain a good accuracy on the normal, unperturbed/non-poisoned test set.  \nnrepair can also improve the overall accuracy of the models, however the effectiveness of the repair tends to decrease when the original accuracy is already high.
\end{tcolorbox}

\textbf{RQ2:}
Table~\ref{tab:summary} can be used to compare the performance of intermediate-layer and last-layer repair on the different scenarios.
For the MNIST models, last-layer repair did not help in improving the overall accuracy. Repairing the dense layer before the output layer using the pattern-based repair helps in increasing the accuracy albeit by a small amount. For the CIFAR10 model, on the other hand, repairing the output layer increases the overall accuracy of the model by 0.16, which is better than intermediate-layer repair (+0.03). 

For the poisoned and adversarial scenarios, on the MNIST models, last-layer repair performed better than intermediate layer-repair on the targeted test sets. Intermediate-layer repair increased the accuracy by 1.81 on the poisoned model and 3.87 on the adversarial model while last-layer repair increased the accuracy by 45.56 on the poisoned model and 10.40 on the adversarial model. For CIFAR10-Pois, intermediate layer repair increases the accuracy by 0.81 while last layer repair improves it by 3.77.  Note that intermediate-layer repair seems to help better in retaining the accuracy on the standard Test, albeit providing smaller improvements on the target sets (detailed results in the Appendix). Furthermore, for CIFAR10-Adv, intermediate layer repair gives better results than last-layer repair (0.34 vs 0.27 respectively).


To summarize, focusing only on an inner layer of the network or just the output layer may not suffice to correct errors in all models and scenarios. We plan to investigate application of repair at more than one layer. Fault localization approaches may help determine the layer/s to focus on for effective repair for a given application. 

\begin{tcolorbox}
\textbf{Answer RQ2:} 
Intermediate-layer repair helped more in improving the overall accuracy of the models (except for CIFAR10) and last-layer repair was more effective in repairing specific failures such as vulnerabilities to poisoned or adversarial inputs (except on CIFAR10 adversarial model). The take away is that there is not a specific type of repair (last-layer or intermediate layer) that works well consistently and different models and failure scenarios may necessitate repair at different layers. 
\end{tcolorbox}

\textbf{RQ3:} 
To understand the overhead introduced by running multiple experts and the combination logic, we conducted experiments on one of the models, MNIST-LQ.
We executed the original model on the test set and compared the inference time with the model produced by a repair at an intermediate layer (i.e., layer 6) and by a repair at the final layer (i.e., layer 8).
Additionally, we measured the inference time for an intermediate layer repair with F1-Filtering (i.e., layer6-F1).
We performed this comparison for all 10,000 inputs in the test set.

\begin{figure}
\begin{center}
\includegraphics[width=0.8\textwidth]{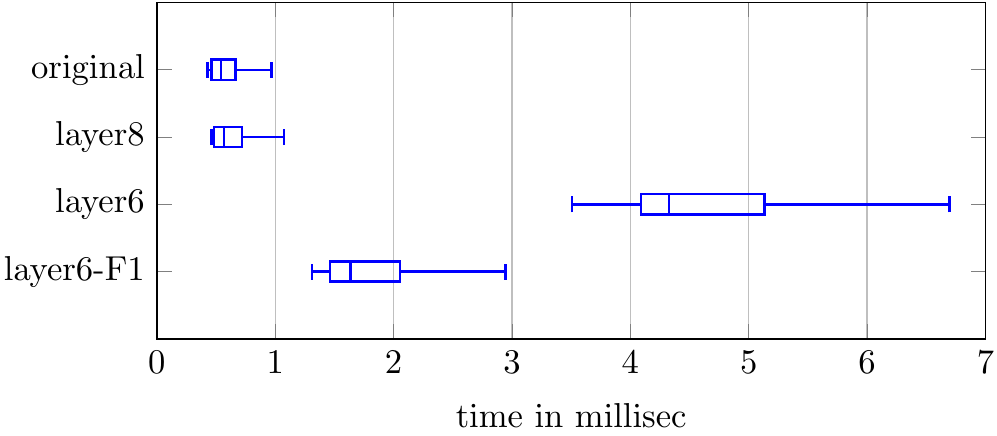}
\end{center}
\caption{Inference Time Comparison (\textit{Naive} Combination Strategy)}
\label{fig:inferencetime}
\end{figure}

The \textit{Merge} combination strategy does not require any expert combination after model execution because this strategy merges the repairs in advance.
Therefore, there is no change with regard to the original model execution except the weight values used in the calculations, and we did not observe any difference in terms of the inference time.
We focus the remaining discussion on the strategies that require the execution of multiple experts.
Our experiments show that the time for the expert combination after model execution (as necessary for \textit{Naive}, \textit{Confidence}, and \textit{Voting} combination strategies) is negligible with around 0.0008 ms and also is similar for all these combination strategies.
The main overhead is introduced by the additional calculations necessary to compute the multiple expert values at each layer.
The box plot in Figure \ref{fig:inferencetime} shows the total time for the model execution for the experts inclusive the time for the \textit{Naive} expert combination.

The repair at the last layer produces an average slowdown (compared to the original model) of 1.0383x.
In contrast, the repair at the intermediate layer produces an average slowdown of 7.7638x.
Therefore, it makes sense to apply some filtering of experts, which do not show good performance on the training set (see F1-score filtering in Section \ref{subsec:expert-combination}).
For this experiment we kept 3 experts (see the plot with \textit{layer6-F1}).
This reduced model produces an average slowdown of only 3.0742x.

\vspace{1em}
\begin{tcolorbox}
\textbf{Answer RQ3:}
The \textit{Merge} combination strategy does not impact the inference time.
All other combination strategies introduce a similar overhead.
While the inference time for the last-layer repair is comparable with the original model, the inference time for an intermediate-layer repair is expensive.
However, it can be significantly reduced with F1 filtering.
\end{tcolorbox}

\subsection{Discussion}

The purpose of our evaluation was to showcase the versatility of \nnrepair in different scenarios. 
The takeaway from the experiments is that there is not a specific type of repair (last-layer or intermediate layer) that works well consistently and different models and failure scenarios may necessitate repair at different layers. In particular, we believe that the intermediate-layer repair holds the most promise for scaling to large networks and we plan to further experiment with the technique in the future. 

Generally, the best repair results are obtained on the poisoning task, where the accuracy can be increased by up to 45\% and 3.7\% on MNIST and CIFAR10, without a need for retraining, which can be expensive in practice. Furthermore, note that we do not assume knowledge of the poison, as our techniques only use information about correct and incorrect classification. In the future, we plan to perform more experiments with different poisoning scenarios.

We were able to obtain modest accuracy improvements on the high-quality models, while for the low-quality models, re-training can achieve better results (see comparison with MODE in the next section). More experimental comparison with retraining and/or fine-tuning the models is needed to further assess the merits of our constraint-based repair.

The gains in the adversarial setting are not very significant for the larger models. In this work, our goal was to demonstrate the feasibility of using localized constraints solving as a generic technique for addressing a wide range of challenges in deep learning. Adversarial attack is only one potential application scenario that is considered. There is a large body of research work on adversarial attacks and we can not claim in any way that we can cover all attacks.

We also note that the efficacy of \nnrepair is evaluated statistically (over the test set) as our method does not provide any formal guarantees. In general, it is difficult to guarantee an improvement of the overall accuracy with formalisms, as there are no formal specifications for the image classification domain. Thus, in practice one builds (trains) a model using a statistical measure of accuracy.

%% file: tables/summary.tex
\begin{table}[!ht]
	\centering
	\caption{Summary of \nnrepair performance on all models. Repair column shows the type of repair, i.e., intermediate or last layer. Increase/decrease in accuracy shown in terms of the difference between the accuracy of the repaired model and the original model on the respective datasets. Accuracy of the original model is shown in brackets (in bold) below each data set. The \emph{Strategy} column shows the combination strategy which work best for each scenario. \textit{ALL} means that all strategies performed equally. \emph{F1-Filter} shows if best results were obtained by turning F1-Filter ON or OFF. The number of experts used are shown in brackets.}
	
\label{tab:summary}
\scalebox{0.9}{
		\begin{tabular}{|l|l|c|c|c|c|c|c|c|}
			\hline
			\textbf{Model} & \textbf{Repair} & \multicolumn{4}{c|}{\textbf{Increase/Decrease in Accuracy}}  & \textbf{Strategy} & \textbf{F1-Filter}\\ \hline
            
                       &  &  \multicolumn{2}{c|}{{\bf Train}}  &  \multicolumn{2}{c|}{{\bf Test}} &&\\
                       &  &  \multicolumn{2}{c|}{{\bf (96.59\%)}}  &  \multicolumn{2}{c|}{{\bf (96.34\%)}} &&\\
            
            MNIST-LQ           &  Interm.&  \multicolumn{2}{c|}{+0.22}  &  \multicolumn{2}{c|}{+0.20} & Votes & ON(3)\\
           
            MNIST-LQ           &  Last& \multicolumn{2}{c|}{+0.00}       &  \multicolumn{2}{c|}{+0.00}  & ALL & ON(0)\\\hline
             
                 &  &  \multicolumn{2}{c|}{{\bf Train}}  &  \multicolumn{2}{c|}{{\bf Test}} &  & \\
                 &  &  \multicolumn{2}{c|}{{\bf (99.81\%)}}  &  \multicolumn{2}{c|}{{\bf (98.89\%)}} &  & \\
            
            MNIST-HQ           &  Interm.&  \multicolumn{2}{c|}{+0.01}  &  \multicolumn{2}{c|}{+0.02} & Merged &  ON(3)\\ 
           
           MNIST-HQ           &  Last&  \multicolumn{2}{c|}{+0.00}  &  \multicolumn{2}{c|}{+0.00} & ALL &  ON(0)\\ \hline
           
             &  &  \multicolumn{2}{c|}{\textbf{P-Train}} & \multicolumn{1}{c|}{\textbf{Test}} & \multicolumn{1}{c|}{\textbf{P-Test}}  &  & \\ 
            &  &  \multicolumn{2}{c|}{\textbf{(98.99\%)}} & \multicolumn{1}{c|}{\textbf{(98.63\%)}} & \multicolumn{1}{c|}{\textbf{(10.38\%)}}  &  & \\ 
           
            MNIST-Pois.      &  Interm.&  \multicolumn{2}{c|}{+0.00}  &  \multicolumn{1}{c|}{-0.01}  &  \multicolumn{1}{c|}{+1.81}  & Votes & ON(2)\\
           
            MNIST-Pois.      &  Last         &  \multicolumn{2}{c|}{-2.60}  &  \multicolumn{1}{c|}{-3.11}  &  \multicolumn{1}{c|}{+45.56}  & Confidence & OFF\\ \hline
           
             &  & \textbf{Train}& \textbf{Adv-Train}  & \textbf{Test} & \textbf{Adv-Test} &  & \\ 
             &  & \textbf{(98.67\%)}& \textbf{(29.92\%)}  & \textbf{(97.87\%)} & \textbf{(28.37\%)} &  & \\ 
           
            MNIST-Adv.   &  Interm.&  -4.35 &  +2.75  &-4.15   & +3.87  & Confidence & ON(9)\\
           
            MNIST-Adv.   &  Last &-3.99  & +11.15   &-3.14   & +10.40  & Merged & ON(10)\\\hline
          &  &  \multicolumn{2}{c|}{{\bf Train}}  &  \multicolumn{2}{c|}{{\bf Test}} &  & \\
         &  &  \multicolumn{2}{c|}{{\bf (87.25\%)}}  &  \multicolumn{2}{c|}{{\bf (81.04\%)}} &  & \\
          
            CIFAR10           &  Interm.&  \multicolumn{2}{c|}{+0.03}  &  \multicolumn{2}{c|}{+0.03} & Merged &  ON(1)\\ 
           
           CIFAR10           &  Last&  \multicolumn{2}{c|}{+0.12}  &  \multicolumn{2}{c|}{+0.16} & ALL &  ON(1)\\ \hline

             &  &  \multicolumn{2}{c|}{\textbf{P-Train}} & \multicolumn{1}{c|}{\textbf{Test}} & \multicolumn{1}{c|}{\textbf{P-Test}}  &  & \\ 
             &  &  \multicolumn{2}{c|}{\textbf{(96.97\%)}} & \multicolumn{1}{c|}{\textbf{(72.26\%)}} & \multicolumn{1}{c|}{\textbf{(15.89\%)}}  &  & \\
           
            CIFAR10-Pois.      &  Interm. &  \multicolumn{2}{c|}{+0.03}  &  \multicolumn{1}{c|}{+0.02}  &  \multicolumn{1}{c|}{+0.81}  & Merged & ON(4) \\
       
            CIFAR10-Pois.      &  Last &  \multicolumn{2}{c|}{-0.89}  &  \multicolumn{1}{c|}{-0.61}  &  \multicolumn{1}{c|}{+3.77}  & Merged & OFF \\ \hline
      
             &  & \textbf{Train}& \textbf{Adv-Train}  & \textbf{Test} & \textbf{Adv-Test} &  & \\ 
            &  & \textbf{(87.25\%)}& \textbf{(34.39\%)}  & \textbf{(81.04\%)} & \textbf{(35.96\%)} &  & \\ 
           
            CIFAR10-Adv.   &  Interm. &  +0.05   & +0.22   &    -0.07 & +0.34   & Merged & ON(10)\\

            CIFAR10-Adv.   &  Last &   -0.25  & +0.37   &  -0.27   &  +0.27  & Merged & ON(10)\\\hline

\hline
  
\end{tabular}
}
\end{table}

%% file: relatedwork.tex
\section{Related Work}
\label{sec:related}


The emphasis of this paper is on neural network repair, where the goal is to ``correct'' the neural network and improve its performance, robustness and security, by using a
small number of labeled inputs. There have been relatively few attempts for repairing a neural network. These neural network repair works can be classified given if re-training is needed and/or if there 
is a first step to prioritize neuron weights to fix. A number of fix patterns and challenges for
neural network repair were collected in \cite{islam20repairing}.

In MODE \cite{ma2018mode}, a neural network is said to be buggy for a specific output label if its
test accuracy is lower than the expectation. This is fixed by selecting features
that are critical for the misbehavior via differential analysis using a subset of training data and then
retraining by selecting inputs from the remaining unused training inputs based on the differential heat map.
We ran MODE on the MNIST models from our study. The results are as follows:

\begin{center}
{
   \centering
    \begin{tabular}{|c|c|}
    \hline
    Model & Test Acc. (\%)\\ 
    \hline
    MNIST-LQ & +0.37\\
    MNIST-HQ & -0.40\\
    \hline
    \end{tabular}
}
\end{center}

\nnrepair has similar performance, i.e., slightly better than MODE on MNIST-HighQuality and slightly worse on MNIST-LowQuality. 
Meanwhile, the re-training procedure in MODE led to varied performances for the repaired model. The results for MODE are the average outcome after 10 runs, none of which improved the accuracy of MNIST-HighQuality. 

Unlike MODE that identifies ill-trained weights or buggy neurons, Apricot \cite{zhang2019apricot} first generates
a set of models from the original neural network with a reduced set of training data 
and at each iteration of the training, Apricot adjusts each weight of the repaired model towards the average 
weight of these reduced models correctly classifying the input while away from the misclassifications.
The approach from \cite{sotoudeh2019correcting} uses constraint solving for
repairing neural networks. It considers a two-dimension
slice of the input space of ACAS Xu and uses SMT constraints to achieve weight changes for correct cases that
are checked against the specification. We found it non-trivial to extend this approach to typically high-dimensional
input space of the image classifiers that we study in this paper.



Typically, a software repair technique (including for neural networks)  employs as a first step \textit{fault localization} to determine the code entities that need to be fixed.
%
DeepFault \cite{eniser2019deepfault} is an approach to spectrum-based fault localization that aims to identify the neurons that are `more' responsible to adversarial behaviours of a neural network.
However, the aim of DeepFault is to generate more adversarial examples, which is the opposite to the repair purpose of our paper. Another related approach, Arachne~\cite{sohn2019search}, uses fault localization
to identify neural weights (connected to the final output layer) to modify, using Particle Swarm 
Optimisation (PSO), for better weights to improve the model's accuracy on some particular label.
As also noted in \cite{sohn2019search}, increasing the prediction accuracy for a particular label
often comes along with the decreasing prediction accuracy of the overall neural network model.

Our \nnrepair work provides a general repair approach which can be applied for improving accuracy, enhancing robustness against adversarial attacks and fixing the backdoor security problems for neural networks. Although previous techniques could be presumably extended to these scenarios, in practice they were only demonstrated for improving the prediction accuracy of the neural network (in MODE and Apricot) or a particular label (in Arachne).

%% file: conclusion.tex
\section{Conclusion and Future Work}

We presented \nnrepair, which uses constraint solving for  intermediate-layer and last-layer repair of neural networks. We demonstrated \nnrepair in three scenarios: improving the \textit{overall accuracy}, fixing security vulnerabilities caused by \textit{data poisoning} and improving the \textit{adversarial robustness} of the networks.

In future work, we plan to 
experiment with different localization techniques and to evaluate our repair on larger networks and different architectures. 
Our method can also be applied to multiple layers but we restricted to single-layer for scalability. One avenue for research is to apply single-layer repair repeatedly or compositionally to handle correcting bugs across multiple layers.


%% file: appendix/appendix.tex
\section{Appendix}

\input{tables/passingtest}

\input{appendix/low}

\input{appendix/high}

\input{appendix/cifar}

\input{appendix/inter-poisoned}
\input{appendix/last-poisoned}

\input{appendix/inter-adv}
\input{appendix/last-adv}

\input{appendix/inter-cifar-poisoned}

\input{appendix/last-cifar-poisoned}

\input{appendix/inter-cifar-adversarial}
\input{appendix/last-cifar-adversarial}

%% file: tables/passingtest.tex
\begin{table}
	\centering
	\caption{Increase in accuracy wrt number of passing tests from normal train-set used to generate the repair. (P-Test for Poisoned models and Ad-Test for Adversarial models). Bold values highlight the highest value for each row. {N/A denotes cases when the symbolic execution did not finish within our timeout.}}
\label{tab:comparison}
		\begin{tabular}{|l|l|r|r|r|r|}
		 \hline
		   \multirow{2}{*}{\textbf{Model}} & \multirow{2}{*}{\textbf{Repair}} & \multicolumn{4}{c|}{\textbf{Passing Tests}}\\	\cline{3-6}

		   	    &  &  \textbf{0} & \textbf{10} & \textbf{50} & \textbf{100}\\	\hline
            MNIST-Pois.    &  Interm.&+1.06&+0.36 &\textbf{+1.81} &+0.55 \\
            MNIST-Pois.    &  Last&   +39.97&+42.08  &\textbf{+45.56}  &+39.09 \\ \hline

            MNIST-Adv.    &  Interm.& \textbf{+3.87} & +2.71&  +3.62& +2.11\\
            MNIST-Adv.   &  Last&   +7.80& \textbf{+10.40}& +9.72 & +9.70\\ \hline

            CIFAR10-Pois.    &  Interm.&  \textbf{+0.81}& +0.41 &  +0.48& N/A\\
            CIFAR10-Pois.    &  Last& \textbf{+3.77}  &  +3.09& +3.68 & +1.48 \\ \hline

            CIFAR10-Adv.    &  Interm.&  \textbf{+0.34} & +1.80&  +1.62& N/A\\
            CIFAR10-Adv.    &  Last&  +0.24&  +0.24&  \textbf{+0.27}& \textbf{+0.27}\\ \hline
           
           \hline
\end{tabular}
\vspace{-5mm}
\end{table}

%% file: appendix/low.tex
\begin{table}
	\centering
	\caption{Performance of \nnrepair on \textbf{MNIST - LowQuality model}. Results are shown for both \textbf{Intermediate Layer and Last Layer Repair}. Cases where an increase in accuracy was observed are in bold. \emph{OFF} and \emph{ON} column shows the accuracy when the F1-Filter was used and not used respectively. \emph{Experts} column shows the classes whose experts were used after F1-Filtering was applied.}
\label{tab:inter-low}
		\begin{tabular}{|c|c|cc|cc|c|cc|cc|}
	\hline
	\textbf{Repair Type} & \multicolumn{5}{c|}{\textbf{Intermediate-Layer}}&  \multicolumn{5}{c|}{\textbf{Last-Layer}}\\ 
	\hline
	\multirow{3}{*}{\textbf{Strategy}} & \multirow{2}{*}{\textbf{Dataset}} & \multicolumn{2}{c|}{\textbf{Train}} & \multicolumn{2}{c|}{\textbf{Test}} & \multirow{2}{*}{\textbf{Dataset}} & \multicolumn{2}{c|}{\textbf{Train}} & \multicolumn{2}{c|}{\textbf{Test}}\\

    &&\multicolumn{2}{c|}{\textbf{(96.59)}} & \multicolumn{2}{c|}{\textbf{(96.34)}} &  & \multicolumn{2}{c|}{\textbf{(96.59)}} & \multicolumn{2}{c|}{\textbf{(96.34)}}\\\cline{2-11}

	 &\multirow{1}{*}{\textbf{Experts}} & \textbf{OFF} & \textbf{ON} & \textbf{OFF} & \textbf{ON}&\multirow{1}{*}{\textbf{Experts}} & \textbf{OFF} & \textbf{ON} & \textbf{OFF} & \textbf{ON}\\	\hline
	\emph{Naive}    &  \multirow{4}{*}{6,8,9}  & 96.56   &   \textbf{96.80}   &   \textbf{96.35}   &   \textbf{96.54} & \multirow{4}{*}{None} &  96.06    & 96.59   &   95.79 & 96.34\\
	\emph{Merged}   &   &  96.56  &   \textbf{96.73}   &   96.30   &   \textbf{96.38}&  &  84.56    & 96.59   &   84.55 & 96.34 \\ 	
	\emph{Confidence} &  & 93.25 &   \textbf{96.80}   &   93.50   &   \textbf{96.54}&  &  91.46    & 96.59   &   91.41 & 96.34\\ 
	\emph{Votes}   &     & 96.57  &   \textbf{96.81}   &   96.34   &   \textbf{96.54}  &  &  95.97    & 96.59   &   95.72 & 96.34\\ \hline
\end{tabular}
\end{table}


%% file: appendix/high.tex
\begin{table}
	\centering
	\caption{Performance of \nnrepair on \textbf{MNIST - HighQuality model}. Results are shown for both \textbf{Intermediate Layer and Last Layer Repair}. Cases where an increase in accuracy was observed are in bold. \emph{OFF} and \emph{ON} column shows the accuracy when the F1-Filter was used and not used respectively. \emph{Experts} column shows the classes whose experts were used after F1-Filtering was applied.}
\label{tab:inter-low}
		\begin{tabular}{|c|c|cc|cc|c|cc|cc|}
	\hline
	\textbf{Repair Type} & \multicolumn{5}{c|}{\textbf{Intermediate-Layer}}&  \multicolumn{5}{c|}{\textbf{Last-Layer}}\\ 
	\hline
	\multirow{3}{*}{\textbf{Strategy}} & \multirow{2}{*}{\textbf{Dataset}} & \multicolumn{2}{c|}{\textbf{Train}} & \multicolumn{2}{c|}{\textbf{Test}} & \multirow{2}{*}{\textbf{Dataset}} & \multicolumn{2}{c|}{\textbf{Train}} & \multicolumn{2}{c|}{\textbf{Test}}\\

    &&\multicolumn{2}{c|}{\textbf{(99.81)}} & \multicolumn{2}{c|}{\textbf{(98.89)}} &  & \multicolumn{2}{c|}{\textbf{(99.81)}} & \multicolumn{2}{c|}{\textbf{(98.89)}}\\\cline{2-11}

	 &\multirow{1}{*}{\textbf{Experts}} & \textbf{OFF} & \textbf{ON} & \textbf{OFF} & \textbf{ON}&\multirow{1}{*}{\textbf{Experts}} & \textbf{OFF} & \textbf{ON} & \textbf{OFF} & \textbf{ON}\\	\hline
	\emph{Naive}    &\multirow{4}{*}{1,6,8}   &99.81    &99.81      &98.89   &\textbf{98.91}   & \multirow{4}{*}{None}   & 99.81    & 99.81   &98.89   &98.89    \\
	\emph{Merged}   &   &\textbf{99.84}    &\textbf{99.82}      &98.88   &\textbf{98.91}   &    & 99.74    & 99.81   &98.82   &98.89    \\	
	\emph{Confidence} &   &99.81    &99.81      &98.89   &98.89  &  & 99.81    & 99.81   &98.89   &98.89    \\
	\emph{Votes}   &   &99.81    &99.81      &98.89   &98.89   &     & 99.81    & 99.81   &98.89   &98.89    \\ \hline
\end{tabular}
\end{table}


%% file: appendix/cifar.tex
\begin{table}
	\centering
	\caption{Performance of \nnrepair on \textbf{CIFAR10 model}. Results are shown for both \textbf{Intermediate Layer and Last Layer Repair}. Cases where an increase in accuracy was observed are in bold. \emph{OFF} and \emph{ON} column shows the accuracy when the F1-Filter was used and not used respectively. \emph{Experts} column shows the classes whose experts were used after F1-Filtering was applied.}
\label{tab:inter-low}
		\begin{tabular}{|c|c|cc|cc|c|cc|cc|}
	\hline
	\textbf{Repair Type} & \multicolumn{5}{c|}{\textbf{Intermediate-Layer}}&  \multicolumn{5}{c|}{\textbf{Last-Layer}}\\ 
	\hline
	\multirow{3}{*}{\textbf{Strategy}} & \multirow{2}{*}{\textbf{Dataset}} & \multicolumn{2}{c|}{\textbf{Train}} & \multicolumn{2}{c|}{\textbf{Test}} & \multirow{2}{*}{\textbf{Dataset}} & \multicolumn{2}{c|}{\textbf{Train}} & \multicolumn{2}{c|}{\textbf{Test}}\\

    &&\multicolumn{2}{c|}{\textbf{(87.25)}} & \multicolumn{2}{c|}{\textbf{(81.04)}} &  & \multicolumn{2}{c|}{\textbf{(87.25)}} & \multicolumn{2}{c|}{\textbf{(81.04)}}\\\cline{2-11}

	 &\multirow{1}{*}{\textbf{Experts}} & \textbf{OFF} & \textbf{ON} & \textbf{OFF} & \textbf{ON}&\multirow{1}{*}{\textbf{Experts}} & \textbf{OFF} & \textbf{ON} & \textbf{OFF} & \textbf{ON}\\	\hline
	\emph{Naive}    &\multirow{4}{*}{9}   &87.23     &87.25 &    81.03&81.04   &\multirow{4}{*}{5} &\textbf{87.35} &\textbf{87.37} &\textbf{81.05} &\textbf{81.20}\\
	\emph{Merged}   & &87.25     &\textbf{87.28} &81.04    &\textbf{81.07}   & &87.01 &\textbf{87.37} &80.94 &\textbf{81.20}\\	
	\emph{Confidence} &  &86.77     &87.25 &80.86    &81.04 & &87.06 &\textbf{87.37} &80.88 &\textbf{81.20}\\ 
	\emph{Votes}   &   & 87.23    &87.25 &\textbf{81.06}    &81.04    & &\textbf{87.36} &\textbf{87.37} &\textbf{81.10} &\textbf{81.20}\\ \hline
\end{tabular}
\end{table}


%% file: appendix/inter-poisoned.tex
\begin{table}
	\centering
	\caption{Performance of \nnrepair using \textbf{Intermediate Layer Repair on MNIST - Poisoned model}. Cases where an increase in accuracy was observed are in bold. \emph{\# Test} shows the number of passing tests used for repair from the normal train-set. \emph{OFF} and \emph{ON} column shows the accuracy when the F1-Filter was used and not used respectively. \emph{Experts} column shows the classes whose experts were used after F1-Filtering was applied.}
\label{tab:inter-poison1}
			\begin{tabular}{|c|c|c|cc|cc|cc|cc|}
			\hline
		
			\multirow{3}{*}{\textbf{\# Test}} &\multirow{3}{*}{\textbf{Strategy}} &\multirow{2}{*}{\textbf{Dataset}} & \multicolumn{2}{c|}{\textbf{Train}} & \multicolumn{2}{c|}{\textbf{Test}} & \multicolumn{2}{c|}{\textbf{P-Test}}\\

		&&& \multicolumn{2}{c|}{\textbf{(98.99)}} & \multicolumn{2}{c|}{\textbf{(98.63)}} & \multicolumn{2}{c|}{\textbf{(10.38)}}\\\cline{3-9}

			&  & \multirow{1}{*}{\textbf{Experts}}  & \textbf{OFF} & \textbf{ON} & \textbf{OFF} & \textbf{ON} & \textbf{OFF} & \textbf{ON} \\	\hline 
	\multirow{4}{*}{\textbf{0}}&\emph{Naive}  & \multirow{5}{*}{7}  &  98.96	&	98.99	&	98.59	&	98.62	&	10.38	        &	10.37	 \\ 
	&\emph{Merged} &   & 98.99	&	98.99	&	98.55	&	98.62	&	\textbf{10.75}	&	10.37	 \\ 
	&\emph{Confidence}& & 97.91	&	98.99	&	97.32	&	98.62	&	\textbf{11.44}	&	10.37     \\
	&\emph{Votes} &  &    98.96	&	98.99	&	98.57	&	98.62	&	\textbf{10.49}	&	10.37	  \\\hline
	
	\multirow{4}{*}{\textbf{10}}&\emph{Naive} & \multirow{5}{*}{0,7,8}   &	98.99	        &	98.99	&	98.62	&	98.62	&	10.38	&	10.37  \\ 
	&\emph{Merged} &   & \textbf{99.00}  &	98.99	        &	98.62	&	98.63	&	\textbf{10.74}	&	\textbf{10.54}  \\ 
	&\emph{Confidence}&     &	98.99	        &	98.99	&	98.62	&	98.62	&	\textbf{10.73}	&	10.37  \\
	&\emph{Votes} &     &	98.99	        &	98.99	&	98.62	&	98.62	&	\textbf{10.58}	&	10.37   \\ \hline
	
	\multirow{4}{*}{\textbf{50}}&	\emph{Naive}  & \multirow{5}{*}{5,7}  &   98.99	&	98.99	&	98.63	&	98.62	&	10.38	&	10.38 \\ 
	&\emph{Merged}      &   &      98.99	&	98.99	&	98.63	&	98.62	&	\textbf{10.79}	&	\textbf{10.73}	\\ 
	&\emph{Confidence}   &   &      98.92	&	98.99	&	98.54	&	98.62	&	\textbf{11.04}	&	\textbf{10.81}	\\
	&\emph{Votes}        &   &      98.99	&	98.99	&	98.63	&	98.62	&	\textbf{10.62}	&	\textbf{12.19}	\\\hline 
	
    \multirow{4}{*}{\textbf{100}}	&\emph{Naive}  &   \multirow{5}{*}{1,5,7} 	&	98.99	&	98.99	&	98.63	&	98.62	&	10.38	&	10.37 \\
	&\emph{Merged}      &      &	98.99	&	98.99	&	98.61	&	98.60	&	\textbf{10.83}	&	\textbf{10.85} \\ 
	&\emph{Confidence}   &      &	98.99	&	98.99	&	98.62	&	98.62	&	\textbf{10.93}	&	\textbf{10.81}  \\
	&\emph{Votes}        &      &	98.99	&	98.99	&	98.63	&	98.62	&	\textbf{10.64}	&	\textbf{10.68} \\ \hline
	
\end{tabular}
\end{table}


%% file: appendix/last-poisoned.tex
\begin{table}
	\centering
	\caption{Performance of \nnrepair using \textbf{Last Layer Repair on MNIST - Poisoned model}. Cases where an increase in accuracy was observed are in bold. \emph{\# Test} shows the number of passing tests used for repair from the normal train-set. F1-Filtering removed all experts (reverting to original model) so results without F1-Filtering are shown.}
\label{tab:last-poison}
		\begin{tabular}{|c|ccc|ccc|ccc|ccc|}
			\hline
			\textbf{Dataset} & \textbf{Train} & \textbf{Test} & \textbf{P-Test} & \textbf{Train} & \textbf{Test} & \textbf{P-Test}  & \textbf{Train} & \textbf{Test}  & \textbf{P-Test}  & \textbf{Train} & \textbf{Test} & \textbf{P-Test} \\ \hline
			\textbf{\# Test} & \multicolumn{3}{c|}{0} & \multicolumn{3}{c|}{10} &  \multicolumn{3}{c|}{50} & \multicolumn{3}{c|}{100}\\	\hline
	\emph{\textbf{Original}}  &98.99&98.63&10.38&98.99&98.63&10.38&98.99&98.63&10.38&98.99&98.63&10.38   \\ \hline
	\emph{Naive}  &98.98&98.62&10.38&98.99&98.63&10.38&98.99&98.63&10.38&98.86&98.49&10.38   \\ 
	\emph{Merged}&95.73&94.88&\textbf{50.35}&95.54&94.52&\textbf{52.46}&96.33&95.47&\textbf{55.94}&93.99&93.75&\textbf{49.47} \\ \emph{Confidence}&95.82&94.97&\textbf{50.35}&95.58&94.56&\textbf{52.46}&96.39&95.52&\textbf{55.94}&96.48&96.11&\textbf{49.47}\\ 
	\emph{Votes}& 98.97&98.60&10.38&98.99&98.62&10.38&98.99&98.62&10.38&98.84&98.46&10.37   \\ 
	\hline
\end{tabular}
\end{table}


%% file: appendix/inter-adv.tex
\begin{table}
	\centering
	\caption{Performance of \nnrepair using \textbf{Intermediate Layer Repair on MNIST - Adversarial model}. Cases where an increase in accuracy was observed are in bold. \emph{\# Test} shows the number of passing tests used for repair from the normal train-set. \emph{OFF} and \emph{ON} column shows the accuracy when the F1-Filter was used and not used respectively. \emph{Experts} column shows the classes whose experts were used after F1-Filtering was applied.}
\label{tab:inter-adv0}
		\begin{tabular}{|c|c|c|cc|cc|cc|cc|}
			\hline
			 \multirow{3}{*}{\textbf{\# Tests}}	& \multirow{3}{*}{\textbf{Strategy}} & \multirow{2}{*}{\textbf{Dataset}} 	&\multicolumn{2}{c|}{\textbf{Train}} &  \multicolumn{2}{c|}{\textbf{Test}}  &  \multicolumn{2}{c|}{\textbf{Ad-Train}}  &  \multicolumn{2}{c|}{\textbf{Ad-Test}} \\
	    
			 &&&\multicolumn{2}{c|}{\textbf{(98.67)}} &  \multicolumn{2}{c|}{\textbf{(97.87)}}  &  \multicolumn{2}{c|}{\textbf{(29.92)}}  &  \multicolumn{2}{c|}{\textbf{(28.37)}} \\\cline{3-11} 
	    
	    &	 	&\multirow{1}{*}{\textbf{Experts}} & \textbf{OFF} & \textbf{ON} & \textbf{OFF}  & \textbf{ON} & \textbf{OFF} & \textbf{ON} & \textbf{OFF} & \textbf{ON} \\	\hline
   \multirow{4}{*}{\textbf{0}}	 &\emph{Naive} &\multirow{4}{*}{All except 2} &   98.65	&	98.33	&	97.81	&	97.66	&	29.73	&	29.55	&	\textbf{28.38}	&	\textbf{28.74}	\\
	&\emph{Merged}&    &98.43	&	98.45	&	97.65	&	97.59	&	\textbf{31.36}	&	\textbf{31.49}	&	\textbf{30.37}	&	\textbf{30.27}		\\ 
	&\emph{Confidence}&   &94.50	&	94.32	&	93.79	&	93.72	&	\textbf{32.51}	&	\textbf{32.67}	&	\textbf{32.08}	&	\textbf{32.24}		\\ 
	&\emph{Votes}&      &98.51	&	98.20	&	97.72	&	97.59	&	\textbf{30.26}	&	\textbf{30.21}	&	\textbf{29.27}	&	\textbf{29.26}		 \\ \hline

    \multirow{4}{*}{\textbf{10}}	&\emph{Naive}&\multirow{4}{*}{All except 2,5}  &  98.50	&	97.91		&	97.68	&	97.27	&	29.36	&	29.41	&		27.79	&	28.30		\\ 
	&\emph{Merged}&    &91.43	&	98.41	&		90.83	&	97.68	&		\textbf{31.79}	&	\textbf{30.67}		&	\textbf{29.84}	&	\textbf{29.28}	\\ 
	&\emph{Confidence}&     &96.48	&	96.24	&	95.78	&	95.61	&	\textbf{31.76}	&	\textbf{31.95}		&	\textbf{31.01}	&	\textbf{31.08}\\ 
	&\emph{Votes}&      & 98.44	&	97.85	&	97.65	&	97.26	&		29.88	&	29.66	&		\textbf{28.54}	&	\textbf{28.66} \\ \hline
    
	\multirow{4}{*}{\textbf{50}} &\emph{Naive}&  \multirow{4}{*}{All except 2} &  98.66	&	98.35	&97.81	&	97.68	&	29.60	&	29.63		&	28.16	&	\textbf{28.50}		\\
	&\emph{Merged}&  &  98.42	&	98.41	&	97.68	&	97.52	&	\textbf{30.17}	&	\textbf{30.35}		&	\textbf{28.82}	&	\textbf{28.80}		\\ 
	&\emph{Confidence}&          &95.30	&	95.19	&	94.65	&	94.63	&	\textbf{32.58}	&	\textbf{32.69}		&	\textbf{31.90}	&	\textbf{31.99} 	 \\ 
	&\emph{Votes}   &   &   98.63	&	98.35	&	97.81	&	97.68	  &	\textbf{30.03}	&	\textbf{30.15}			&	\textbf{28.66}	&	\textbf{28.90}		 \\ \hline

	\multirow{4}{*}{\textbf{100}}&\emph{Naive}& \multirow{4}{*}{All except 2}  &   98.65	&	98.51	&	97.81	&	97.74	&	29.30	&	29.55		&	27.66	&	27.86		\\ 
	&\emph{Merged}&    &     98.31	&	98.34		&	97.61	&	97.56	&	\textbf{30.15}	&	\textbf{30.35}	&		\textbf{28.64}	&	\textbf{28.41}	\\ 
	&\emph{Confidence}&         & 95.65	&	95.63	&	94.90	&	94.88	&	\textbf{31.38}	&	\textbf{32.02}	&	\textbf{29.97}	&	\textbf{30.48}	 \\
	&\emph{Votes}&          &   98.59	&	98.47		&	97.80	&	97.76	&	29.82	&	\textbf{30.02}	&		28.25	&	28.31	 \\ \hline
\end{tabular}
\end{table}


%% file: appendix/last-adv.tex
\begin{table}
	\centering
	\caption{Performance of \nnrepair using \textbf{Last Layer Repair on MNIST - Adversarial model}. Cases where an increase in accuracy was observed are in bold. \emph{\# Test} shows the number of passing tests used for repair from the normal train-set. \emph{OFF} and \emph{ON} column shows the accuracy when the F1-Filter was used and not used respectively. \emph{Experts} column shows the classes whose experts were used after F1-Filtering was applied.}
\label{tab:last-adv0}
		\begin{tabular}{|c|c|c|cc|cc|cc|cc|}
			\hline
	     \multirow{3}{*}{\textbf{\# Tests}}	& \multirow{3}{*}{\textbf{Strategy}} & \multirow{2}{*}{\textbf{Dataset}} 	&\multicolumn{2}{c|}{\textbf{Train}} &  \multicolumn{2}{c|}{\textbf{Test}}  &  \multicolumn{2}{c|}{\textbf{Ad-Train}}  &  \multicolumn{2}{c|}{\textbf{Ad-Test)}} \\
	    
	    & &	&\multicolumn{2}{c|}{\textbf{(98.67)}} &  \multicolumn{2}{c|}{\textbf{(97.87)}}  &  \multicolumn{2}{c|}{\textbf{(29.92)}}  &  \multicolumn{2}{c|}{\textbf{(28.37)}} \\\cline{3-11}

	    &	 	&\multirow{1}{*}{\textbf{Experts}} & \textbf{OFF} & \textbf{ON} & \textbf{OFF}  & \textbf{ON} & \textbf{OFF} & \textbf{ON} & \textbf{OFF} & \textbf{ON} \\	\hline
	
   \multirow{4}{*}{\textbf{0}}	 &\emph{Naive} &\multirow{4}{*}{All}&98.67&98.67&97.87&97.87&29.92&29.92&\textbf{28.38}&\textbf{28.38}   \\ 
	&\emph{Merged}&&93.30&93.30&93.59&93.59&\textbf{38.15}&\textbf{38.15}&\textbf{36.17}&\textbf{36.17}  \\ 
	&\emph{Confidence}&&93.30&93.30&93.59&93.59&\textbf{38.15}&\textbf{38.15}&\textbf{36.17}&\textbf{36.17}\\ 
	&\emph{Votes}&&98.67&98.67&97.87&97.87&\textbf{29.96}&\textbf{29.96}&\textbf{28.38}&\textbf{28.38}  \\ \hline
	
    \multirow{4}{*}{\textbf{10}}   & \emph{Naive} &\multirow{4}{*}{All}  &98.67&98.67&97.87&97.87&\textbf{29.96}&\textbf{29.96}&\textbf{28.39}&\textbf{28.39}   \\ 
    &\emph{Merged}&&94.68&94.68&94.73&94.73&\textbf{41.07}&\textbf{41.07}&\textbf{38.77}&\textbf{38.77}  \\ 
	&\emph{Confidence}&&94.70&94.70&94.74&94.74&\textbf{40.94}&\textbf{40.94}&\textbf{38.70}&\textbf{38.70}\\ 
	&\emph{Votes}&& \textbf{98.69}&\textbf{98.69}&\textbf{97.94}&\textbf{97.94}&\textbf{30.16}&\textbf{30.16}&\textbf{28.47}&\textbf{28.47}  \\ \hline	
    
    \multirow{4}{*}{\textbf{50}}  & \emph{Naive} &\multirow{4}{*}{All}  &98.66&98.66&97.87&97.87&\textbf{30.09}&\textbf{30.09}&\textbf{28.54}&\textbf{28.54}   \\ 
	&\emph{Merged}&&95.33&95.33&95.26&95.26&\textbf{40.11}&\textbf{40.11}&\textbf{38.09}&\textbf{38.09}  \\ 
    &\emph{Confidence}&&95.49&95.49&95.39&95.39&\textbf{40.07}&\textbf{40.07}&\textbf{38.01}&\textbf{38.01}\\ 
	&\emph{Votes}&&98.66&98.66&97.82&97.82&\textbf{30.22}&\textbf{30.22}&\textbf{28.56}&\textbf{28.56}   \\ \hline
	
    \multirow{4}{*}{\textbf{100}}   &	\emph{Naive} &\multirow{4}{*}{All except 2}  &98.63&98.34&97.83&97.52&\textbf{30.50}&\textbf{31.73}&\textbf{29.39}&\textbf{30.38}   \\ 
	&\emph{Merged}&&95.33&95.67&94.75&95.35&\textbf{38.96}&\textbf{40.32}&\textbf{36.76}&\textbf{38.07}  \\ 
	&\emph{Confidence}&&95.94&96.62&95.63&96.07&\textbf{38.37}&\textbf{39.56}&\textbf{36.02}&\textbf{37.01}\\ 
	&\emph{Votes}&& 98.62&98.30&97.78&97.47&\textbf{30.67}&\textbf{31.99}&\textbf{29.62}&\textbf{30.57}  \\ \hline
\end{tabular}
\end{table}

%% file: appendix/inter-cifar-poisoned.tex
\begin{table}
	\centering
	\caption{Performance of \nnrepair using \textbf{Intermediate Layer Repair on CIFAR10 - Poisoned model}. Cases where an increase in accuracy was observed are in bold. \emph{\# Test} shows the number of passing tests used for repair from the normal train-set. \emph{OFF} and \emph{ON} column shows the accuracy when the F1-Filter was used and not used respectively. \emph{Experts} column shows the classes whose experts were used after F1-Filtering was applied.}
\label{tab:inter-poison1}
			\begin{tabular}{|c|c|c|cc|cc|cc|cc|}
			\hline
		
			\multirow{3}{*}{\textbf{\# Test}} &\multirow{3}{*}{\textbf{Strategy}} &\multirow{2}{*}{\textbf{Dataset}} & \multicolumn{2}{c|}{\textbf{(Train)}} & \multicolumn{2}{c|}{\textbf{Test}} & \multicolumn{2}{c|}{\textbf{P-Test}}\\ 
			 & & & \multicolumn{2}{c|}{\textbf{(96.97)}} & \multicolumn{2}{c|}{\textbf{(72.26)}} & \multicolumn{2}{c|}{\textbf{(15.89)}}\\\cline{3-9}

			&  & \multirow{1}{*}{\textbf{Experts}}  & \textbf{OFF} & \textbf{ON} & \textbf{OFF} & \textbf{ON} & \textbf{OFF} & \textbf{ON} \\	\hline 

	\multirow{4}{*}{\textbf{0}}	 &\emph{Naive}  &\multirow{4}{*}{2,5,6,7} &\textbf{96.98}  	&96.97	&\textbf{72.78}		&72.26		&15.89		&\textbf{15.90}			 \\ 
    &   \emph{Merged}&&96.97 &\textbf{97.00}  	&72.26		&\textbf{72.28}		&15.89		&\textbf{16.70}							 \\ 
	&  \emph{Confidence}&&96.58 &96.97  	&\textbf{72.48}		&72.26		&\textbf{16.44}		&\textbf{15.90}				 \\  
	&  \emph{Votes}&&\textbf{96.98}&96.97 	&\textbf{72.30}			&72.26	&\textbf{16.20}		&\textbf{15.90}			 		 \\ \hline

	\multirow{4}{*}{\textbf{10}}	 &\emph{Naive}  &\multirow{4}{*}{7} &\textbf{96.98}  	&\textbf{96.98}		&\textbf{72.27}		&\textbf{72.35}		&15.89		&15.72			 \\ 
    &   \emph{Merged}&&96.97 &96.97  	&72.26		&72.26		&15.89		&15.89							 \\ 
	&  \emph{Confidence}&&96.68 &\textbf{96.98}  	&\textbf{72.43}		&\textbf{72.35}		&\textbf{16.30}		&15.72				 \\  
	&  \emph{Votes}&&\textbf{96.98}&\textbf{96.98} 	&\textbf{72.28}		&\textbf{72.35}		&\textbf{16.14}		&15.72			 		 \\ \hline

	\multirow{4}{*}{\textbf{50}}	 &\emph{Naive}  &\multirow{4}{*}{7} &96.97 	&\textbf{96.98}		&72.24		&\textbf{72.31}		&15.89		&15.71			 \\ 
    &   \emph{Merged}&&96.97 &96.97  	&72.26		&72.26		&15.89		&15.89							 \\ 
	&  \emph{Confidence}&&\textbf{96.99} &\textbf{96.98}  	&\textbf{72.45}		&\textbf{72.31}		&\textbf{16.37}		&15.71				 \\  
	&  \emph{Votes}&&96.96&\textbf{96.98} 	&72.25				&\textbf{72.31}	&\textbf{16.01}	&15.71			 		 \\ \hline


\end{tabular}
\end{table}


%% file: appendix/last-cifar-poisoned.tex
\begin{table}
	\centering
	\caption{Performance of \nnrepair using \textbf{Last Layer Repair on CIFAR10 - Poisoned model}. Cases where an increase in accuracy was observed are in bold. \emph{\# Test} shows the number of passing tests used for repair from the normal train-set. \emph{OFF} and \emph{ON} column shows the accuracy when the F1-Filter was used and not used respectively. \emph{Experts} column shows the classes whose experts were used after F1-Filtering was applied.}
\label{tab:inter-poison1}
			\begin{tabular}{|c|c|c|cc|cc|cc|cc|}
			\hline
		
			\multirow{3}{*}{\textbf{\# Test}} &\multirow{3}{*}{\textbf{Strategy}} &\multirow{2}{*}{\textbf{Dataset}} & \multicolumn{2}{c|}{\textbf{Train)}} & \multicolumn{2}{c|}{\textbf{Test}} & \multicolumn{2}{c|}{\textbf{P-Test}}\\ 
			 & & & \multicolumn{2}{c|}{\textbf{(96.97)}} & \multicolumn{2}{c|}{\textbf{(72.26)}} & \multicolumn{2}{c|}{\textbf{(15.89)}}\\\cline{3-9}

			&  & \multirow{1}{*}{\textbf{Experts}}  & \textbf{OFF} & \textbf{ON} & \textbf{OFF} & \textbf{ON} & \textbf{OFF} & \textbf{ON} \\	\hline 
	\multirow{4}{*}{\textbf{0}}&\emph{Naive}  & \multirow{4}{*}{5}  &   	96.97	&	\textbf{97.00}	&	\textbf{72.27}	&	72.25	&	15.89	&	\textbf{15.95}		 \\ 
	&\emph{Merged} & &   96.08	&	\textbf{97.00}	&	71.65	&	72.25	&	\textbf{19.66}	&	\textbf{15.95}		 	 \\ 
	&\emph{Confidence}& & 96.08	&	\textbf{97.00}	&	71.65	&	72.25	&	\textbf{19.66}	&	\textbf{15.95}		      \\
	&\emph{Votes} & & 96.97	&	\textbf{97.00}	&	\textbf{72.27}	&	72.25	&	15.89	&	\textbf{15.95}		 	  \\\hline
	
	\multirow{4}{*}{\textbf{10}}&\emph{Naive}  & \multirow{4}{*}{5}  & 96.94	&	\textbf{97.00}	&	72.25	&	72.25	&	15.87	&	\textbf{15.95}		 \\ 
	&\emph{Merged} &   &  	95.98	&	\textbf{97.00}	&	71.52	&	72.25	&	\textbf{18.95}	&	\textbf{15.95}		 	 \\ 
	&\emph{Confidence}& & 96.10	&	\textbf{97.00}	&	71.66	&	72.25	&	\textbf{18.98}	&	\textbf{15.95}		      \\
	&\emph{Votes} & & 96.94	&	\textbf{97.00}	&	72.25	&	72.25	&	15.87	&	\textbf{15.95}
		 	  \\\hline

	\multirow{4}{*}{\textbf{50}}&\emph{Naive}  & \multirow{4}{*}{None}  &  	96.97	&	96.97	&	72.26	&	72.26	&	15.89	&	15.89		 \\ 
	&\emph{Merged} &   &  	96.12	&	96.97	&	71.79	&	72.26	&	\textbf{19.57}	&	15.89	 	 \\ 
	&\emph{Confidence}& &  96.12	&	96.97	&	71.79	&	72.26	&	\textbf{19.57}	&	15.89		      \\
	&\emph{Votes} &  &  96.97	&	96.97	&	72.26	&	72.26	&	15.89	&	15.89
 	  \\\hline
	
    \multirow{4}{*}{\textbf{100}}&\emph{Naive}  & \multirow{4}{*}{5}  &  	96.92	&	\textbf{96.99}	&	72.20	&	72.16	&	15.89	&	\textbf{16.04}		 \\ 
	&\emph{Merged} &   &  	95.97	&	\textbf{96.99}	&	71.61	&	72.16	&	\textbf{17.37}	&	\textbf{16.04}		 	 \\ 
	&\emph{Confidence}& &  96.15	&	\textbf{96.99}	&	71.70	&	72.16	&	\textbf{17.37}	&	\textbf{16.04}	      \\
	&\emph{Votes} &  &  	96.92	&	\textbf{96.99}	&	72.20	&	72.16	&	15.88	&	\textbf{16.04}		 	  \\\hline
	
\end{tabular}
\end{table}


%% file: appendix/inter-cifar-adversarial.tex
\begin{table}
	\centering
	\caption{Performance of \nnrepair using \textbf{Intermediate Layer Repair on CIFAR10 - Adversarial model}. Cases where an increase in accuracy was observed are in bold. \emph{\# Test} shows the number of passing tests used for repair from the normal train-set. \emph{OFF} and \emph{ON} column shows the accuracy when the F1-Filter was used and not used respectively. \emph{Experts} column shows the classes whose experts were used after F1-Filtering was applied.}
\label{tab:last-cifar0}
		\begin{tabular}{|c|c|c|cc|cc|cc|cc|}
			\hline
	     \multirow{3}{*}{\textbf{\# Tests}}	& \multirow{3}{*}{\textbf{Strategy}} & \multirow{2}{*}{\textbf{Dataset}} 	&\multicolumn{2}{c|}{\textbf{Train}} &  \multicolumn{2}{c|}{\textbf{Test}}  &  \multicolumn{2}{c|}{\textbf{Ad-Train}}  &  \multicolumn{2}{c|}{\textbf{Ad-Test}} \\
	    &	& 	&\multicolumn{2}{c|}{\textbf{(87.25)}} &  \multicolumn{2}{c|}{\textbf{(81.04)}}  &  \multicolumn{2}{c|}{\textbf{(34.00)}}  &  \multicolumn{2}{c|}{\textbf{(35.96)}} \\\cline{3-11} 

	    &	 	&\multirow{1}{*}{\textbf{Experts}} & \textbf{ON} & \textbf{ON} & \textbf{OFF}  & \textbf{ON} & \textbf{OFF} & \textbf{ON} & \textbf{OFF} & \textbf{ON} \\	\hline
	
	\multirow{4}{*}{\textbf{0}}	 &\emph{Naive}  &\multirow{4}{*}{ALL} &  87.24	&	87.25	&81.01		&81.04		&\textbf{34.03}		&\textbf{34.39}		&\textbf{36.04} 	&35.96	 \\ 
    &   \emph{Merged}&&87.25 &\textbf{87.30}  	&81.03		&80.97	&\textbf{34.02}		&\textbf{34.61}		&\textbf{35.97}		&\textbf{36.30} 		 \\ 
	&  \emph{Confidence}&&86.57 &87.25  	&80.59		&81.04		&\textbf{35.91}		&\textbf{34.39}		&\textbf{37.96}		&35.96	 \\  
	&  \emph{Votes}&&87.24&87.25  &80.93		&81.04		&\textbf{34.03}		&\textbf{34.39}		&\textbf{35.97} 		&35.96 \\ \hline

	\multirow{4}{*}{\textbf{10}}	 &\emph{Naive}  &\multirow{4}{*}{ALL} & 87.23 	&87.23		&81.03		&81.03		&\textbf{34.03}		&\textbf{34.03}		&\textbf{36.02} 	&\textbf{36.02}	 \\ 
    &   \emph{Merged}&&87.25 & 87.25 	&81.03		&81.03		&\textbf{34.03}		&\textbf{34.03}		&35.96		&35.96 		 \\ 
	&  \emph{Confidence}&&86.64 & 86.64 	&80.68		&80.68		&\textbf{35.71}	&\textbf{35.71}		&\textbf{37.76}		&\textbf{37.76}	 \\  
	&  \emph{Votes}&&87.21&87.21  	&80.97		&80.97		&33.98		&33.98		&35.96	&35.96 		 \\ \hline

	\multirow{4}{*}{\textbf{50}}	 &\emph{Naive}  &\multirow{4}{*}{ALL} &87.24  	&87.24		&81.03		&81.03		&\textbf{34.03}		&\textbf{34.03}		&\textbf{36.02} 	&\textbf{36.02}	 \\ 
    &   \emph{Merged}&&\textbf{87.26} &\textbf{87.26}  	&81.04		&81.04		&\textbf{34.02}		&\textbf{34.02}		&35.94		&35.94 		 \\ 
	&  \emph{Confidence}&&86.74 &86.74  	&80.71		&80.71		&\textbf{35.54}		&\textbf{35.54}		&\textbf{37.58}		&\textbf{37.58}	 \\  
	&  \emph{Votes}&&87.24&87.24  	&80.94		&80.94		&\textbf{34.06}		&\textbf{34.06}		&\textbf{36.01}		&\textbf{36.01} 		 \\ \hline

	
\end{tabular}
\end{table}

%% file: appendix/last-cifar-adversarial.tex
\begin{table}
	\centering
	\caption{Performance of \nnrepair using \textbf{Last Layer Repair on CIFAR10 - Adversarial model}. Cases where an increase in accuracy was observed are in bold. \emph{\# Test} shows the number of passing tests used for repair from the normal train-set. \emph{OFF} and \emph{ON} column shows the accuracy when the F1-Filter was used and not used respectively. \emph{Experts} column shows the classes whose experts were used after F1-Filtering was applied.}
\label{tab:last-cifar0}
		\begin{tabular}{|c|c|c|cc|cc|cc|cc|}
			\hline
	     \multirow{3}{*}{\textbf{\# Tests}}	& \multirow{3}{*}{\textbf{Strategy}} & \multirow{2}{*}{\textbf{Dataset}} 	&\multicolumn{2}{c|}{\textbf{Train}} &  \multicolumn{2}{c|}{\textbf{Test}}  &  \multicolumn{2}{c|}{\textbf{Ad-Train}}  &  \multicolumn{2}{c|}{\textbf{Ad-Test}} \\
	    &	& 	&\multicolumn{2}{c|}{\textbf{(87.25)}} &  \multicolumn{2}{c|}{\textbf{(81.04)}}  &  \multicolumn{2}{c|}{\textbf{(34.00)}}  &  \multicolumn{2}{c|}{\textbf{(35.96)}} \\\cline{3-11} 

	    &	 	&\multirow{1}{*}{\textbf{Experts}} & \textbf{ON} & \textbf{ON} & \textbf{OFF}  & \textbf{ON} & \textbf{OFF} & \textbf{ON} & \textbf{OFF} & \textbf{ON} \\	\hline
	
	\multirow{4}{*}{\textbf{0}}	 &\emph{Naive}  &\multirow{4}{*}{All} &  87.25	&	87.25	&	81.04	&	81.04	&	34.00	&	34.00	&	35.96	&	35.96 \\ 
    &   \emph{Merged}&& 86.32	&	86.32	&	80.15	&80.15	&	33.96&	33.96	&	\textbf{36.20}	&\textbf{36.20} \\ 
	&  \emph{Confidence}&&  86.21	&	86.21	&	80.11	&	80.11	&	\textbf{34.02}	&	\textbf{34.02}	&	\textbf{36.18}	&\textbf{36.18}\\ 
	&  \emph{Votes}&& 87.25	&	87.25	&	81.04	&	81.04	&	34.00&	34.00	&	35.96	&	35.96  \\ \hline

     \multirow{4}{*}{\textbf{10}}	 &	\emph{Naive} &\multirow{4}{*}{All} & 87.25	&	87.25	&	81.03	&81.03	&	33.97	&	33.97	&	\textbf{35.98}	&	\textbf{35.98} \\ 
    &  \emph{Merged}&&  86.32	&	86.32	&	80.15&	80.15	&	33.96	&	33.96	&	\textbf{36.20}	&	\textbf{36.20} \\
	&  \emph{Confidence}&&  86.30	&	86.30	&	80.15	&	80.15	&	33.89	&	33.89	&	\textbf{36.17}&	\textbf{36.17} \\ 
	&  \emph{Votes}&&   87.25	&	87.25	&	81.03	&	81.03	&	33.97	&	33.97	&	\textbf{35.98}	&	\textbf{35.98}\\ \hline

	\multirow{4}{*}{\textbf{50}}	 &\emph{Naive}  &\multirow{4}{*}{All}& 87.25	&	87.25	&	81.04	&	81.04	&	34.00	&	34.00	&	35.93	&	35.93  \\ 
	&  \emph{Merged}&& 87.00	&	87.00	&	80.77	&	80.77	&	\textbf{34.37}	&	\textbf{34.37}	&	\textbf{36.23}	&	\textbf{36.23} \\
	&  \emph{Confidence}&&    86.85	&	86.85	&	80.64	&	80.64	&	\textbf{34.16}	&	\textbf{34.16}	&	\textbf{36.21}	&	\textbf{36.21} \\ 
	&  \emph{Votes}&&  87.25	&	87.25	&	81.04	&	81.04	&	34.00&	34.00	&	35.93	&	35.93 \\ \hline

	\multirow{4}{*}{\textbf{100}}	 &\emph{Naive} &\multirow{4}{*}{All} & 87.25	&	87.25	&	\textbf{81.05}	&	\textbf{81.05}	&	\textbf{34.01}	&	\textbf{34.01}	&	35.94	&	35.94  \\ 
	&  \emph{Merged}&& 87.00	&	87.00	&	80.77	&	80.77	&	\textbf{34.37}	&	\textbf{34.37}	&	\textbf{36.23}&	\textbf{36.23} \\ 
	&  \emph{Confidence}&&  86.99	&	86.99	&	80.78	&	80.78	&	\textbf{34.39}	&	\textbf{34.39}	&	\textbf{36.21}	&	\textbf{36.21}\\ 
	&  \emph{Votes}&&  87.25	&	87.25	&	\textbf{81.05}	&	\textbf{81.05}	&	\textbf{34.01}	&	\textbf{34.01}	&	35.94&	35.94 \\ \hline
	
\end{tabular}
\end{table}